\documentclass[twoside,11pt]{article}
\usepackage{blindtext}
\usepackage[abbrvbib, preprint]{jmlr2e}
 
\usepackage{amsmath,amssymb,graphicx,mathrsfs,bbm,url}
\usepackage{hyperref}
    \hypersetup{colorlinks=true,citecolor=blue,linkcolor=blue,urlcolor=black}

\usepackage[table]{xcolor}

\usepackage{float}
\usepackage{graphicx}
\usepackage{subcaption}
\usepackage{caption}
\usepackage{lscape}
\usepackage{booktabs}
\usepackage{multirow}

\usepackage{xparse}

\usepackage{comment}
\usepackage{marginnote}
\definecolor{MidnightBlue}{RGB}{25,25,112}
\definecolor{MidnightBlueComplementingGreen}{RGB}{25,112,25}
\definecolor{MidnightBlueComplementingPurple}{RGB}{112,25,112}
\definecolor{MidnightBlueComplementingRed}{RGB}{112,25,69}
\definecolor{deepjunglegreen}{rgb}{0.0, 0.29, 0.29}
\definecolor{applegreen}{rgb}{0.55, 0.71, 0.0}
\definecolor{WowColor}{rgb}{.75,0,.75}
\definecolor{MildlyAlarming}{rgb}{0.85,0.25,0.1}
\definecolor{SubtleColor}{rgb}{0,0,.50}
\definecolor{SubtleColor2}{rgb}{0.6,0.21,.50}
\newcounter{margincounter}

\NewDocumentCommand{\Anastasis}{mo}{
    \IfValueF{#2}{
                        {{\scriptsize
                            \textcolor{deepjunglegreen}{ 
                            \textbf{A:}
                            \textit{{#1}}
                            }
                        }}
        }
    \IfValueT{#2}{
                        \marginnote{{\scriptsize
                            \textcolor{deepjunglegreen}{ 
                            \textbf{A:}
                            \textit{{#1}}
                            }
                        }}
        }
                    }
\NewDocumentCommand{\Ruiyang}{mo}{
    \IfValueF{#2}{
                        {{\scriptsize
                            \textcolor{applegreen}{ 
                            \textbf{R:}
                            \textit{{#1}}
                            }
                        }}
        }
    \IfValueT{#2}{
                        \marginnote{{\scriptsize
                            \textcolor{applegreen}{ 
                            \textbf{R:}
                            \textit{{#1}}
                            }
                        }}
        }
                    }

\newcommand{\eqdef}{\ensuremath{\stackrel{\mbox{\upshape\tiny def.}}{=}}}

\NewDocumentCommand{\Exp}{mo}{
    \operatorname{exp}_{{#1}}
    \IfValueT{#2}{({#2})}
}

\NewDocumentCommand{\Log}{mo}{
    \operatorname{log}_{{#1}}
    \IfValueT{#2}{({#2})}
}

\newtheorem{thm}{Theorem}[section]

\newtheorem{lem}[thm]{Lemma}
\newtheorem{cor}[thm]{Corollary}

\ShortHeadings{Capacity Bounds for Hyperbolic Neural Network Representations of Latent Tree Structures}{Kratsios, Hong, Saez de Ocariz Borde}
\firstpageno{1}

\begin{document}

\title{Capacity Bounds for Hyperbolic Neural Network Representations of Latent Tree Structures}

\author{\name Anastasis Kratsios\thanks{Corresponding author.}
        \email kratsioa@mcmaster.ca \\
        \addr Department of Mathematics, McMaster University and the Vector Institute
        \AND
        \name Ruiyang Hong \email hongr5@mcmaster.ca\\
        \addr Department of Mathematics, McMaster University and the Vector Institute
        \AND
        \name Haitz Sáez de Oc\'{a}riz Borde \email haitz@oxfordrobotics.institute\\
        \addr Oxford Robotics Institute, University of Oxford}

\maketitle

\begin{abstract}
We study the representation capacity of deep hyperbolic neural networks (HNNs) with a ReLU activation function.  We establish the first proof that HNNs can $\varepsilon$-isometrically embed any finite weighted tree into a hyperbolic space of dimension $d$ at least equal to $2$ with prescribed sectional curvature $\kappa<0$, for any $\varepsilon> 1$ (where $\varepsilon=1$ being optimal).  We establish rigorous upper bounds for the network complexity on an HNN implementing the embedding.  We find that the network complexity of HNN implementing the graph representation is independent of the representation fidelity/distortion.  We contrast this result against our lower bounds on distortion which any ReLU multi-layer perceptron (MLP) must exert when embedding a tree with $L>2^d$ leaves into a $d$-dimensional Euclidean space, which we show at least $\Omega(L^{1/d})$; independently of the depth, width, and (possibly discontinuous) activation function defining the MLP.
\end{abstract}

\noindent \textbf{Keywords:} Generalization Bounds, Graph Neural Networks, Digital Hardware, Discrete Geometry, Metric Embeddings, Discrete Optimal Transport, Concentration of Measure.

\noindent \textbf{MSC (2022)}: 68T07, 30L05, 68R12, 05C05.

\section{Introduction}
\label{s:Introduction}

Trees are one of the most important hierarchical data structures in computer science, whose structure can be exploited to yield highly efficient algorithms.  For example, leveraging the tree's hierarchical structure to maximize parameter search efficiency, as in branch-and-bound algorithms \cite{land2010automatic} or depth-first searches \cite{korf1985depth}.  Consequentially, algorithms designed for trees and algorithms which map more general structures into trees, e.g.\ \cite{fuchs1980visible}, have become a cornerstone of computer science and its related areas.  Nevertheless, it is known that the flat Euclidean geometry of $\mathbb{R}^d$ is fundamentally different from the expansive geometry of trees which makes them difficult to embed into low dimensional Euclidean space with low distortion \cite{Bourgain_1986_IJM__EmbeddingMetricSpacesSuperreflexivBanachSpaces,Matouvek_1999_IJM__EmbeddingsTreesUCBanSpaces,Gupta_2000_ACM__QuantitiativefinitedimembeddingsTrees}.  This high distortion can be problematic for downstream tasks relying on Euclidean representations of such trees.  This fundamental impasse in representing large trees with Euclidean space has sparked the search for non-Euclidean representation spaces whose geometry is tree-like, thus allowing for low-dimensional representation of arbitrary trees.  One such family of representation spaces are the hyperbolic spaces $\mathbb{H}^d$, $d \ge 2$, which have recently gained traction in machine learning.

\begin{figure}[hpt!]%[H]%
    \centering
    \begin{subfigure}[t]{.45\textwidth}
        \centering
        \includegraphics[width=1\linewidth]{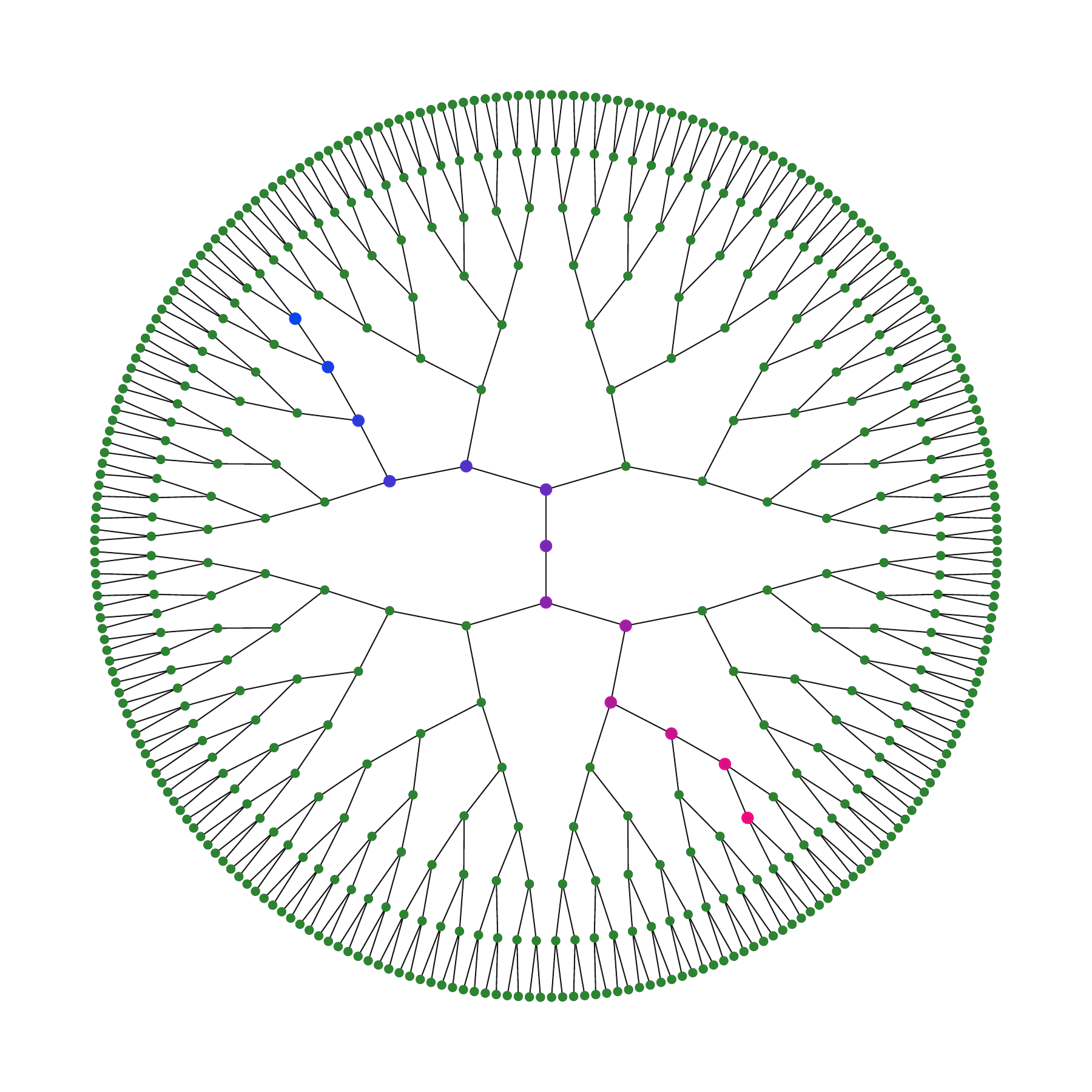}
        \caption{\label{fig:tree} Binary Tree (green) with a coloured minimal length curve in blue-red. }  
    \end{subfigure}
    \centering
    \begin{subfigure}[t]{0.45\textwidth}
        \centering
        \includegraphics[width=.9\linewidth]{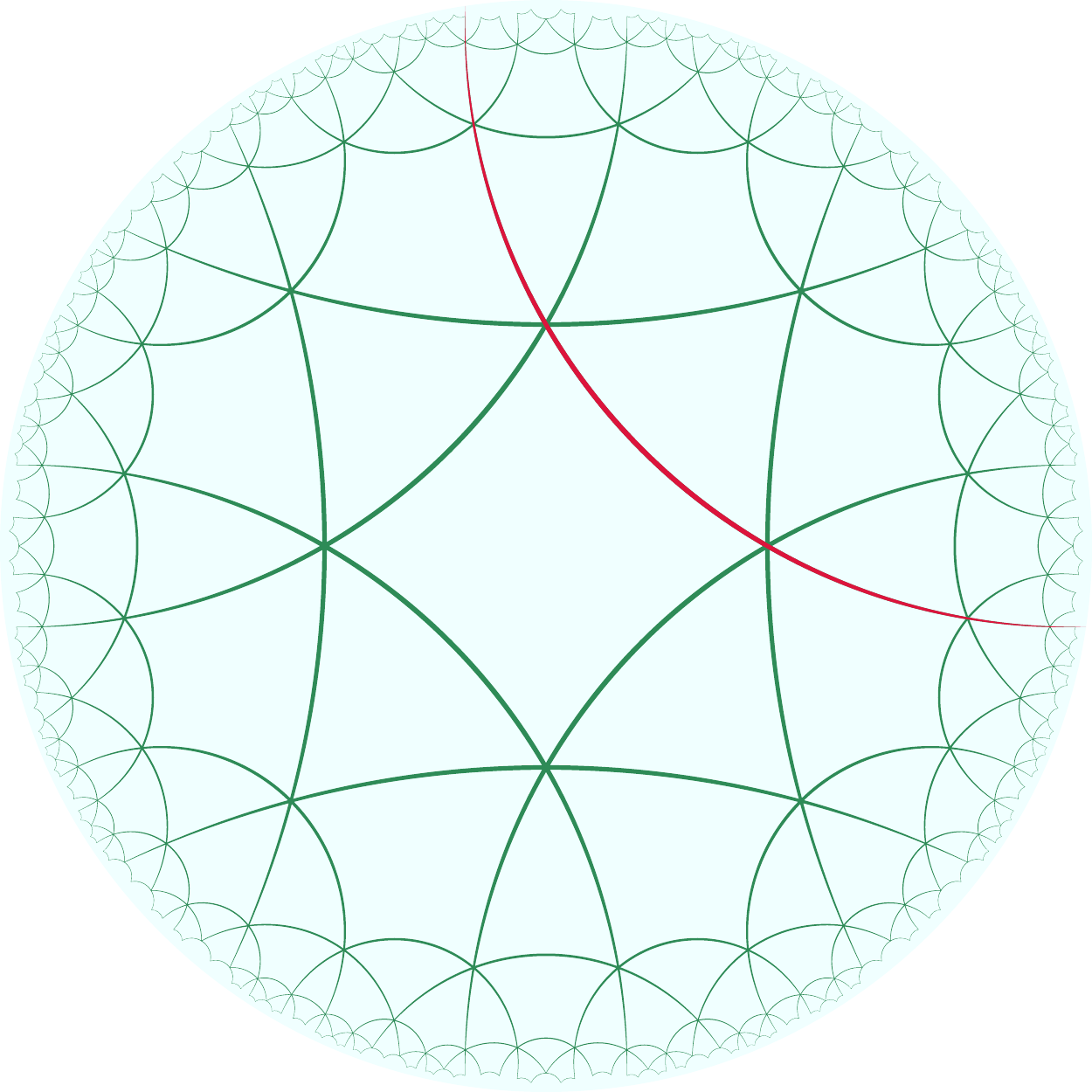}
        \caption{\label{fig:hyperbolic} The two-dimensional hyperbolic space $\mathbb{H}^2$ with minimal length curves in red.}
    \end{subfigure}
    \caption[1]{Minimal length curves in the hyperbolic space\protect\footnotemark (right) expand outwards exponentially quickly just as the number of nodes double exponentially rapidly in a tree (right) as one travels away from the origin/root.}
    \label{fig:Tree_VS_Hyperbolic}
\end{figure}

Leveraging hyperbolic data representations with a (latent) tree-like structure has proven significantly more effective than their traditional Euclidean counterparts.  Machine learning examples include learning linguistic hierarchies \cite{Mikel_1989_MemAmerMathSoc__UniversalRTrees}, natural language processing \cite{ganea2018hyperbolic,zhu2020hypertext}, recommender systems \cite{vinh2020hyperml,skopek2020mixed}, low-dimensional representations of large tree-like graphs \cite{ganea2018hyperbolic2,law2020ultrahyperbolic,kochurov2020hyperbolic,zhu2020graph,bachmann2020constant,sonthalia2020tree}, knowledge graph representations \cite{chami2020low}, network science \cite{papadopoulos2014network,keller2020hydra}, communication \cite{kleinberg2007geographic}, deep reinforcement learning \cite{cetin2023hyperbolic}, and numerous other recent applications.

These results have motivated deep learning on hyperbolic spaces, of which the hyperbolic neural networks (HNNs) \cite{ganea2018hyperbolic}, and their several variants \cite{gulcehre2018hyperbolic,chami2019hyperbolic,shimizu2021hyperbolic,zhang2021hyperbolic} have assumed the role of the flagship deep learning model.  
This has led HNNs to become integral in several deep learning-power hyperbolic learning algorithms and have also fuelled applications ranging from natural language processing \cite{ganea2018hyperbolic,dhingra2018embedding,tay2018hyperbolic,liu2019hyperbolic,zhu2020hypertext}, to latent graph inference for downstream graph neural network (GNN) optimization \cite{kazi2022differentiable,de2022latent}.  Furthermore, the simple structure of HNNs makes them amenable to mathematical analysis, similarly to multi-layer perceptrons (MLPs) in classical deep learning.  This has led to the establishment of their approximation capabilities in \cite{KratsiosBilokopytov_2020_NeurIPS__FirstApproximation,KratsiosPapon_2022_JMLR__UATGDL}.

The central motivation behind HNNs is that they are believed to be better suited to representing data with latent tree-like, or hierarchical, structure than their classical $\mathbb{R}^n$-valued counterparts; e.g.\ MLPs, CNNs, GNNs, or Transformers, since the geometry of hyperbolic space $\mathbb{H}^d$, $d\ge 2$, is most similar to the geometry of trees than classical Euclidean space, see Figure~\ref{fig:Tree_VS_Hyperbolic}.  These intuitions are often fueled by classical embedding results in computer science~\cite{sarkar2011low}, metric embedding theory~\cite{Bonk_Schramm_2000_GAFA__EmbeddingGromovHyperbolicSpaces}, and the undeniable success of countless algorithms leveraging hyperbolic geometry \cite{papadopoulos2012popularity,papadopoulos2014network,nickel2017poincare,balazevic2019multi,sonthalia2020tree,keller2020hydra} for representation learning.  Nevertheless, the representation potential of HNNs, in representing data with latent hierarchies, currently only rests on strong experimental evidence, expert intuition rooted in deep results from hyperbolic geometry \cite{Gromov1981HyperbolicManifolds,Gromov1987HyperbolicGroups,Bonk_Schramm_2000_GAFA__EmbeddingGromovHyperbolicSpaces}.

\footnotetext{In this illustration, we use the Poincar\'{e} model of the hyperbolic space and not the hyperboloid model used in the manuscript and in most software packages; e.g.~\cite{BoumalMishraAbsilSepulchre_2014_JMLR__ManOpt,TownsendKoepWichwald_2016_JMLR__Pymanopt}.}

In this paper, we examine the problem of Euclidean-vs-hyperbolic representation learning when a latent hierarchy structures the data.
We justify this common belief by first showing that HNNs can $\varepsilon$-isometrically embed any pointcloud with a latent weighted tree structure for any $\varepsilon>0$.  In contrast, such an embedding cannot exist in any Euclidean space.  We show that the HNNs implementing these $\varepsilon$-embeddings are relatively small by deriving upper bounds on their depth, width, and number of trainable parameters sufficient for achieving any desired representation capacity.  We find that HNNs only require $\widetilde{\mathcal{O}}(N^2)$ trainable parameters to embed any $N$-point pointcloud in $n$-dimensional Euclidean space, with a latent tree structure, into the $2$-dimensional hyperbolic plane.  

We then return to the problem of Euclidean-vs-hyperbolic representation under a latent hierarchical structure, by proving that any MLP cannot faithfully embed them into a low-dimensional Euclidean space, thus proving that HNNs are superior to MLPs for representing tree-like structures.  We do so by showing that any MLP, regardless of its depth, width, or number of trainable parameters, cannot embed a pointcloud with latent tree structure, with $L>2^d$ leaves, into the $d$-dimensional Euclidean space with distortion less than $\Omega(L^{1/d})$.  We consider the distortion of an embedding as in the classical computer science literature, e.g.\ \cite{LinialLondonRavinovich_Combinatorica_1995__Embeddings,Bartal_FOCS_1996__ProbApproxMetricEmbeddinsAlgoImplications,Gupta_2000_ACM__QuantitiativefinitedimembeddingsTrees}; a formal definition will be given in the main text.

\paragraph{Outline}
The rest of this paper is organized as follows.
Section~\ref{s:Preliminaries} introduces the necessary terminologies for hyperbolic neural networks, such as the geometry of hyperbolic spaces and the formal structure of HNNs. 
Section~\ref{s:RepresentationProblemFormulation} formalizes latent tree structures and then numerically formalizes the representation learning problem.  

Our main graph representation learning results are given in Section~\ref{thm:Main}.
In Section~\ref{s:LowerBounds}, we first derive lower bounds for the best possible distortion achievable by an MLP representation of a latent tree. 
We show that MLPs cannot embed any large tree in a small Euclidean space, irrespective of how many wide hidden layers the network uses and irrespective of which, possibly discontinuous, non-linearity is used to define the MLP.  
In Section~\ref{s:UpperBounds}, we show that HNNs can represent any pointcloud with a latent tree structure to arbitrary precision.  Furthermore, the depth, width, and number of trainable parameters defining the HNN are independent of the representation fidelity/distortion.  Our theory is validated experimentally in Section~\ref{s:Experiments}. 
The analysis and proofs of our main results are contained in Section~\ref{s:Proof}.
We draw our conclusions in Section~\ref{s:Conclusion}.

\section{The Hyperbolic Neural Network Model}
\label{s:Preliminaries}

Throughout this paper, we consider hyperbolic neural networks (HNNs) with the $\operatorname{ReLU}(t)\eqdef \max\{0,t\}$ (Rectified Linear Unit) activation function/non-linearity, mapping into hyperbolic representation spaces $\mathbb{H}^d_{\kappa}$.  This section rigorously introduced the HNN architecture.  This requires a brief overview of hyperbolic spaces, which we do now.
\subsection{The Geometry of the Hyperbolic Spaces\texorpdfstring{ $\mathbb{H}^d_{\kappa}$}{}}
\label{s:Preliminaries__ss:Poincaremodel}

Fix a positive integer $d$ and a \textit{(sectional) curvature parameter} $\kappa<0$.  The (hyperboloid model for the real) hyperbolic $d$-space of constant sectional curvature $\kappa$, denoted by $\mathbb{H}^d_{\kappa}$, consists of all points $x\in \mathbb{R}^{1+d}$ satisfying 
\[
        1+ \sum_{i=1}^d\, x_i^2  = x_{d+1}^2
    \mbox{ and }
        x_{d+1} > 0 
\]
where the distance between any pair of point $x,y\in \mathbb{H}^d$ is given by
\[
        d_{\kappa}(x,y)
    \eqdef 
        \frac1{\sqrt{|\kappa|}}
        \,
        \operatorname{cosh}\Big(
                x_{d+1}y_{d+1}
            -
                \sum_{i=1}^d\,x_iy_i
        \Big)
.
\]
It can be shown that $\mathbb{H}^d_{\kappa}$ is a simply connected\footnote{See \citep[Section 6.4]{Jost_2017_Book__RiemannianGeometryGeometricAnalysis}.} smooth manifold \citep[pages 92-93]{BrisonHaefliger_Book_1999__NPCMetricSpaces} and that the metric $d_{\kappa}$ on any such $\mathbb{H}^d_{\kappa}$ measures the length of the shortest curve joining any two points on $\mathbb{H}^d_{\kappa}$ where length is quantified in the infinitesimal Riemannian sense\footnote{See \citep[page 20]{Jost_2017_Book__RiemannianGeometryGeometricAnalysis}.} \citep[Propositions 6.17 (1) and 6.18]{BrisonHaefliger_Book_1999__NPCMetricSpaces}.  This means that, by the Cartan-Hadamard Theorem \citep[Corollary 6.9.1]{Jost_2017_Book__RiemannianGeometryGeometricAnalysis}, for every $x\in \mathbb{H}^d_{\kappa}$ there is a map $\Exp{x}:T_x(\mathbb{H}^d_{\kappa})\cong \mathbb{R}^d\rightarrow \mathbb{H}^d_{\kappa}$ which puts $\mathbb{R}^d$ into bijection with $\mathbb{H}^d$ in a smooth manner with smooth inverse.  

\begin{remark}
\label{rem:lightnotation}
Often the metric $d_{\kappa}$ is not relevant for a given statement, and only the manifold structure of $\mathbb{H}^d_{\kappa}$ matters, or the choice of $\kappa$ is clear from the context.  In these instances, we write $\mathbb{H}^d$ in place of $\mathbb{H}^d_{\kappa}$ to keep our notation light.
\end{remark}

For any point $x\in \mathbb{R}^d$, we identify with the $d$-dimensional affine subspace $T_x(\mathbb{H}^d)$ of $\mathbb{R}^{d+1}$ lying tangent to $\mathbb{H}^d$ at $x$.  For any $x\in \mathbb{H}^d$, the tangent space $T_x(\mathbb{H}^d)$ is the $d$-dimensional affine subspace of $\mathbb{R}^{d+1}$ consisting of all $y\in \mathbb{R}^{d+1}$ satisfying 
\[
        y_{d+1}x_{d+1}
    =
        \sum_{i=1}^d\,y_ix_i
    .
\]
The tangent space at $\boldsymbol{1}_n$ plays an especially important role since its elements can be conveniently identified with $\mathbb{R}^d$.  This is because $x\in T_{{\boldsymbol{1}_n}}(\mathbb{H}^n)$ only if it is of the form $x=(x_1,\dots,x_n,1)$, for some $x_1,\dots,x_n\in \mathbb{R}$.  Thus,
\begin{equation}
\label{eq:formal_identifications}
        (x_1,\dots,x_n,1) \overset{\pi_n}{\rightarrow} (x_1,\dots,x_n) 
    \mbox{ and }
      (x_1,\dots,x_n) \overset{\iota_n}{\rightarrow} (x_1,\dots,x_n,1)
\end{equation}
identify $T_{{\boldsymbol{1}_n}}(\mathbb{H}^n)$ with $\mathbb{R}^n$.

The map $\Exp{x}$ can be explicitly described as the map which sends any ``initial velocity vector'' $v\in \mathbb{R}^d$ lying tangent to $x\in \mathbb{H}^d$ to the unique point in $\mathbb{H}^d$ which one would arrive at by travelling optimally thereon.  Here, optimally means along the unique minimal length curve in $\mathbb{H}^d_{-1}$, illustrated in Figure~\ref{fig:hyperbolic} \footnote{In the Euclidean space $\mathbb{R}^d$ these are simply straight lines.}.  
The (affine) tangent spaces $T_x(\mathbb{H}^d)$ and $T_y(\mathbb{H}^d)$ about any two points $x$ and $y$ in $\mathbb{H}^d$ are identified by ``sliding'' $T_x(\mathbb{H}^d)$ towards $T_y(\mathbb{H}^d)$ in parallel across the minimal unique length curve joining $x$ to $y$ in $\mathbb{H}^d$.  This ``sliding'' operation, called \textit{parallel transport}, is formalized by the linear isomorphism%
\footnote{In general, parallel transport is path dependant.  However, since we only consider minimal length (geodesic) curves joining points on $\mathbb{H}^d_{\kappa}$, and there is only one such choice by the Cartan-Hadamard theorem; then there is no ambiguity in the notation/terminology in our case.} %
$P_{x\mapsto b}:T_{x}(\mathbb{H}^n)\rightarrow T_b(\mathbb{H}^n)$ given for any $u\in T_{x}(\mathbb{H}^n)$ by%
\footnote{In \cite{KratsiosPapon_2022_JMLR__UATGDL}, the authors note that the identifications $T_c(\mathbb{H}^m)$ with $\mathbb{R}^m$ are all made implicitly.  However, here, we underscore each component of the HNN pipeline by making each identification processed by any computer completely explicit.}
\begin{equation}
\label{eq:classical_Mobius_HNN_Layers__ParallelTransport}
    P_{x\mapsto b}:\,
        u
    \mapsto 
            u 
        -
            \frac{
                \langle 
                        \Log{x}[b]
                    ,
                        u
                \rangle_x
            }{
                d_{-1}^2(x,b)
            }
            \,
            \big(
                    \log_x(b)
                +
                    \log_b(x)
            \big)
,
\end{equation}
where the map $\Log{x}:\mathbb{R}^d\rightarrow\mathbb{H}^d$ is defined for any $y\in \mathbb{H}^d$ by
\[
        \Log{x}: 
        y
    \mapsto 
            \frac{
                d_{-1}(x,y)
                (y- \langle x|y\rangle_M x)
            }{
                \|
                    y- \langle x|y\rangle_M x
                \|_2
            }
    ,
\]
and where $\|\cdot\|_2$ is the usual Euclidean norm on $\mathbb{R}^{d+1}$.  Typically parallel transport must be approximated numerically, e.g.~\cite{GuiguiPennec_2022_FOCM__LatterSchemsParallelTransport}, but this is not so for $\mathbb{H}^d_{-1}$.

The hyperbolic space is particularly convenient, amongst negatively curved Riemannian manifolds, since the map $\Exp{x}$ is available in closed-form.  For any $x\in \mathbb{H}^d_{\kappa}$ and $v\in \mathbb{R}^d$, $\Exp{x}(v)$ is given by: $\Exp{x}(v)=x$ if $v=0$, otherwise
\[
        \Exp{x}:
        v
    \mapsto
            \operatorname{cosh}\big(
                \sqrt{
                    \langle v| v\rangle_M
                }
            \big)
        +
            \operatorname{sinh}\big(
                \sqrt{
                    \langle v| v\rangle_M
                }
            \big)
            \,
            \frac{v}{\langle v| v\rangle_M}
\]
where $\langle u| v\rangle_M \eqdef -u_{d+1}v_{d+1} + \sum_{i=1}^d\,u_iv_i$ for any $u,v\in \mathbb{R}^d$; see \citep[page 94]{BrisonHaefliger_Book_1999__NPCMetricSpaces}.
Furthermore, the inverse of $\Exp{x}$ is $\Log{x}$.  

We note that these operations are implemented in most standard geometric machine learning software, e.g.~\cite{BoumalMishraAbsilSepulchre_2014_JMLR__ManOpt,TownsendKoepWichwald_2016_JMLR__Pymanopt,miolane2020geomstats}.  Also, $\mathbb{H}_{\kappa}^d$ and $\mathbb{H}_{-1}^d$ are diffeomorphic by the Cartan-Hadamard Theorem \citep[Corollary 6.9.1]{Jost_2017_Book__RiemannianGeometryGeometricAnalysis}.  Therefore, for all $\kappa<0$, it suffices to consider the ``standard'' exponential map $\Exp{x}$ in the particular case where $\kappa=-1$ to map encode in $\mathbb{R}^d$ into $\mathbb{H}^d$ and visa-versa via $\Log{x}$.

\subsection{The Hyperbolic Neural Network Model}
\label{s:Preliminaries__ss:HyperbolicNNs}

\begin{figure}[hpt!]
    \centering
        \includegraphics[width=1\linewidth]{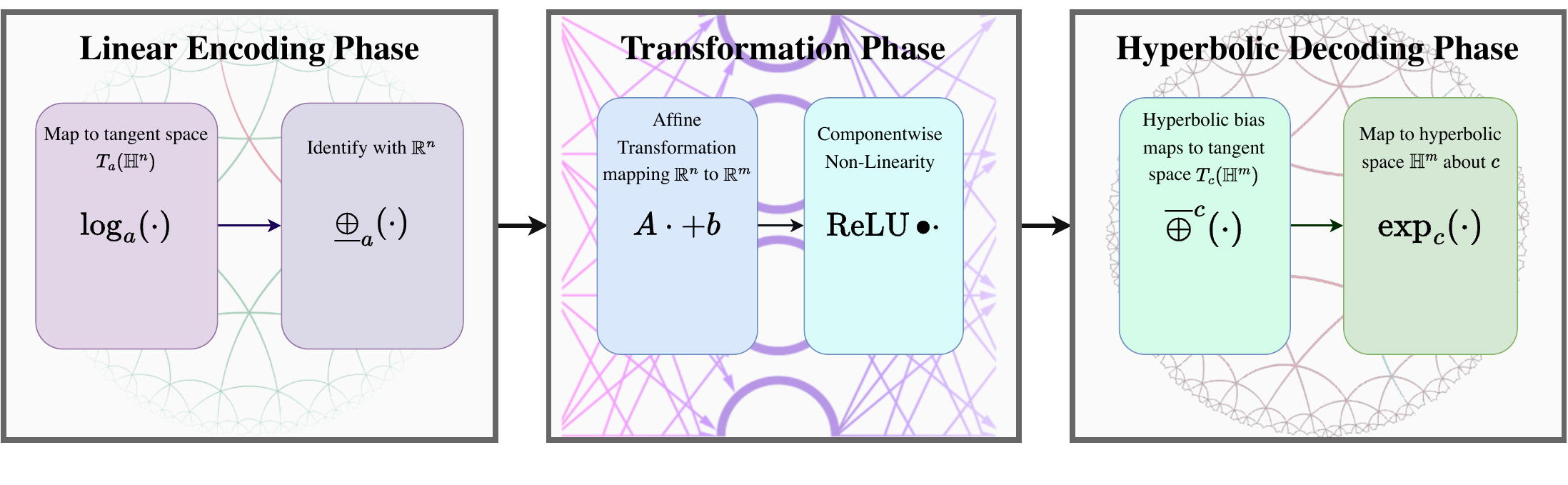}
        \caption{\textbf{Workflow of an HNN layer:}  First inputs in a hyperbolic space $\mathbb{H}^n$ are mapped to a vector in $\mathbb{R}^n$ in the ``encoding phase''.  Next, they are transformed to ``deep features'' in $\mathbb{R}^m$ by a standard MLP layer in the ``transformation phase''.  Finally, the ``hyperbolic decoding phase'' applies a hyperbolic bias, at which point the ``deep features with hyperbolic bias'' are decoded, thus producing an output in the hyperbolic space $\mathbb{H}^m$.}  
    \label{fig:HNN_Layer__Workflow}
\end{figure}

We now overview the hyperbolic neural network model studied in this paper.  The considered HNN model contains the hyperbolic neural networks of \cite{ganea2018hyperbolic}, from the deep approximation theory \cite{KratsiosPapon_2022_JMLR__UATGDL}, and the latent graph inference \cite{de2022latent} literatures as sub-models.  The workflow of the hyperbolic neural network's layer, by which it processes data on the hyperbolic space $\mathbb{H}^n$, is summarized in Figure~\ref{fig:HNN_Layer__Workflow}.

HNNs function similarly to standard MLPs, which generate predictions from any input by sequentially applying affine maps interspersed with non-affine non-linearities, typically via component-wise activation functions.  Instead of leveraging affine maps, which are otherwise suited to the vectorial geometry of $\mathbb{R}^d$, HNNs are built using analogues of the affine maps for Euclidean spaces, which are suited to the geometry of $\mathbb{H}^2$.  Thus, the analogues of the linear layers with component-wise $\operatorname{ReLU}(t)\eqdef \max\{0,t\}$ activation function, mapping $\mathbb{H}^n$ to $\mathbb{H}^m$ for $n,m\in \mathbb{N}_+$, are thus given by
\begin{equation}
\label{eq:classical_Mobius_HNN_Layers__ActivatedLinear}
        x
    \mapsto 
        \Exp{{\boldsymbol{1}}_n}[
            \operatorname{ReLU}\bullet (A \Log{{\boldsymbol{1}}_m}(x))
        ]
\end{equation}
where $\bullet$ denotes component-wise composition and $A$ is an $m\times n$ matrix, and\footnote{Were we have the distinguished ``origin'' point $0$ in the Poincar\'{e} disc model for $\mathbb{H}^n_{\kappa}$ used in \citep[Definition 3.2]{ganea2018hyperbolic} with its corresponding point in the hyperboloid model for $\mathbb{H}^n_{\kappa}$ using the isometry between these two spaces given on \citep[page 86]{BrisonHaefliger_Book_1999__NPCMetricSpaces}.} ${\boldsymbol{1}}_n\in \mathbb{H}^n$.  Thus, as discussed in~\citep[Theorem 4 and Lemma 6]{ganea2018hyperbolic}, without the ``hyperbolic bias'' term, the elementary layers making up HNNs can be viewed as elementary MLP layers conjugated by the maps $\Exp{{\boldsymbol{1}}_n}$ and $\Log{{\boldsymbol{1}}_m}$.  In this case, $\Exp{{\boldsymbol{1}}_n}$ and $\Log{{\boldsymbol{1}}_m}$ serve simply to respectively encode and decode the hyperbolic features into vectorial data, which can be processed by standard software.

The analogues of the addition of a bias term were initially formalized in \cite{ganea2018hyperbolic} using the so-called gyro-vector addition and multiplication operators; see \cite{Vermeer_2005_TopAppl__GeometricInterpretationAdditionGyrovector}, which roughly states that a bias $b\in \mathbb{H}^n$ can be added to any $x\in \mathbb{H}^n$ by ``shifting by'' $b$ along minimal length curves in $\mathbb{H}^n$ using $\Exp{b}$.  Informally,
\begin{equation}
\label{eq:classical_Mobius_HNN_Layers__Bias}
        y
    \mapsto 
        \Exp{b}[P_{{\boldsymbol{1}}_n\mapsto b}\circ \Log{0}[y]]
\end{equation}
where $P_{{\boldsymbol{1}}_n\mapsto b}:T_{{\boldsymbol{1}}_n}(\mathbb{H}^n)\rightarrow T_b(\mathbb{H}^n)$ linearly identifies the tangent space at ${\boldsymbol{1}}_n\eqdef$ $(0,\dots,0,1)$ with that at $b$ by travelling along the unique minimal length curve in $\mathbb{H}^n_{-1}$, defined by $d_{-1}$, connecting ${\boldsymbol{1}}_n$ to $b$.  The interpretation of~\eqref{eq:classical_Mobius_HNN_Layers__Bias} as the addition of a bias term dates back at least to early developments in geometric machine learning in \cite{Pennec_2006_JMIV__IntrinsicStatsManifolds,MeyerBonnabelSepulchre_2011_JMLR__Regression_Matrices,Fletcher_2013_IJCV__GeodesicRegressionManifolds}.  The basic idea is that in the Euclidean space, the analogues of the $\Exp{x}$ and $\Log{x}$ are simply addition and subtraction, respectively.

Here, we consider a generalization of the elementary HNN layers of \cite{ganea2018hyperbolic}, used in \citep[Corollaries 23 and 24]{KratsiosPapon_2022_JMLR__UATGDL} to constructing universal deep learning models capable of approximating continuous functions\footnote{Not intepolators.} between any two hyperbolic spaces.  The key difference here is that they, and we, also allowed for a ``Euclidean bias'' to be added in together with the hyperbolic bias computed by~\eqref{eq:classical_Mobius_HNN_Layers__Bias}.  Similar HNN layers are also used in the contemporary latent graph inference literature~\cite{de2022latent,kazi2022differentiable}.  Incorporating this Euclidean bias, with the elementary maps~\eqref{eq:classical_Mobius_HNN_Layers__Bias}, the hyperbolic biases~\eqref{eq:classical_Mobius_HNN_Layers__ActivatedLinear}, and the formal identifications~\eqref{eq:formal_identifications} we obtain our elementary \textit{hyperbolic} layers $\mathcal{L}_{a,b,c,A}:\mathbb{H}^{n}\rightarrow \mathbb{H}^{m}$ maps given by
\begin{equation}
\label{eq:generalized_HNN_Layers}
        \mathcal{L}_{a,b,c,A}
        :
            x
        \mapsto 
            \Exp{c}\biggl(
                \overline{\oplus}^c
                \Big(
                    \operatorname{ReLU}\bullet
                    \big(
                        A
                        \,
                        \underline{\oplus}_a
                            \circ 
                        \Log{a}[x]
                    +
                        b
                    \big)
                \Big)
            \biggr)
\end{equation}
where \textit{weight matrix} $A$ is an $m\times n$ matrix, a \textit{Euclidean bias} $b\in \mathbb{R}^n$, and the incorporation of \textit{hyperbolic biases} $a\in \mathbb{H}^n$ and $b\in \mathbb{H}^m$ are defined by
\footnote{The notation $\overline{\oplus}^b$ and $\underline{\oplus}_a$ is intentionally similar to the gyrovector-based ``hyperbolic bias translation'' operation in \citep[Equation (28)]{ganea2018hyperbolic} to emphasize the similarity between these operations.}
\[
\begin{aligned}
            \overline{\oplus}^{c}
    \eqdef 
    &
        P_{{\boldsymbol{1}_m\mapsto c}}
        \circ 
        \iota_m
    : \mathbb{R}^m\rightarrow T_c(\mathbb{H}^m)
    \\
        \underline{\oplus}_{a}
    \eqdef 
    &
        \pi_n
            \circ 
        P_{a\mapsto \boldsymbol{1}_{n}}
    : T_a(\mathbb{H}^n) \rightarrow \mathbb{R}^n
.
\end{aligned}
\]
The number of \textit{trainable parameters} defining any hyperbolic layer $\mathcal{L}_{a,b,c,A}$ is
\begin{equation}
\label{eq:parameter_count_layer}
        \operatorname{Par}(\mathcal{L}_{a,b,c,A})
    \eqdef 
        \|A\|_0 + \|a\|_0+\|b\|_0 + \|c\|_0
\end{equation}
where $\|\cdot\|_0$ counts the number of non-zero entries in a matrix or vector.

We work with data represented as vectors in $\mathbb{R}^n$ with a latent tree structure.  
Similarly to \cite{de2022latent,kazi2022differentiable}, these can be \textit{encoded} as hyperbolic features, making them compatible with standard HNN pipelines, using $\operatorname{exp}_{{\boldsymbol{1}_n}}$ as a \textit{feature map}.  Any such feature map is regular in that it preserves the approximation capabilities of any downstream deep learning model, see \citep[Corollary 3.16]{KratsiosBilokopytov_2020_NeurIPS__FirstApproximation} for details.

\begin{definition}[Hyperbolic Neural Networks]
\label{defn:Hyperbolic_NNs}
Let $n,d\in \mathbb{N}_+$.  A function $f:\mathbb{R}^n\rightarrow \mathbb{H}^d$ is called a hyperbolic neural network (HNN) if it admits the iterative representation: for any $x\in \mathbb{R}^n$
\begin{align}
\notag
    % Readout/Final Layer
        f(x)
    &
    =
        \Exp{c^{(I+1)}}\Big(
            \overline{\oplus}^{c^{(I+1)}}
                \big(
                    A^{(I+1)}
                    \,
                    (
                            \underline{\oplus}_{
                                c^{(I)}
                            }
                                \circ 
                            \Log{
                                c^{(I)}
                            }[x^{(I)}]
                        +
                            b^{(I+1)}
                    )
                \big)
        \Big)
\\
\notag
    % Hidden Layers
        x^{(i)}
    & 
    =
        \mathcal{L}_{
            c^{(i-1)}
        ,b^{(i)},c^{(i)},A^{(i)}}(
            x^{(i-1)}
        )
    \qquad\qquad\qquad\qquad\qquad\qquad
        \mbox{for } i=1,\dots,I
\\
\notag
    %% Feature Representation
        x^{(0)} 
    & 
    =
        \Exp{c^{(0)}}(\overline{\oplus}^{c^{(0)}}\,x)
\end{align}
where $I\in \mathbb{N}_+$, $n=d_0,\dots,d_{I+2}=d\in \mathbb{N}_+$, and for $i=1,\dots,I+1$, $A^{(i)}$ is a $d_{i+1}\times d_i$ matrix, $b^{(i)}\in \mathbb{R}^{d_i}$, $c^{(i)}\in \mathbb{H}^{d_{i+1}}\subset \mathbb{R}^{d_{i+1}+1}$, and $c^{(0)}\in \mathbb{H}^{n}\subset \mathbb{R}^{n+1}$.
\end{definition}
In the notation of Definition~\ref{defn:Hyperbolic_NNs}, the integer $I+1$ is called the \textit{depth} of the HNN $f$ and the \textit{width} of $f$ is $\max_{i=0,\dots,I+1}\, d_i$.  Similarly to~\eqref{eq:parameter_count_layer}, the total number of \textit{trainable parameters} defining the HNN $f$, denoted by $\operatorname{Par}(f)$, is tallied by
\[
        \operatorname{Par}(f)
    \eqdef 
        \|c^{(0)}\|_0
        +
        \sum_{i=1}^{I+1}
        \,
            \|A^{(i)}\|_0 + \|b^{(i)}\|_0 + \|c^{(i)}\|_0
.
\]
Note that, since the hyperbolic bias $c^{(i)}$ is shared between any two subsequent layers, in the notation of~\eqref{eq:parameter_count_layer} $\alpha=c^{(i-1)}$ and $c=c^{(i)}$ for any $i=1,\dots,I+1$, then we do not \textit{double count} these parameters.

\section{Representing Learning with Latent Tree Structures}
\label{s:RepresentationProblemFormulation}

We now formally define latent tree structures.  These capture the actual hierarchical structures between points in a pointcloud.  We then formalize what it means to represent those latent tree structures in an ideally low-dimensional representation space with little distortion.  During this formalization process we will recall some key terminologies pertaining to trees.

\subsection{Latent Tree Structure}
\label{s:Preliminaries__ss:LatentTrees}

We now formalize the notation of a \textit{latent tree structure} between members of a pointclouds a vector space $\mathbb{R}^n$.  We draw from ideas in clustering, where the relationship between pairs of points is not captured by reflected by their configuration in Euclidean space but rather through some unobservable latent distance/structure.  
Standard examples from the clustering literature include the Mahalanobis distance \cite{xiang2008learning}, the Minkowski distance or $\ell^{\infty}$ distances \cite{singh2013k}, which are implemented in standard software~\cite{achtert2008elki,de2012pattern}, and many others; e.g.~\cite{ye2017fast,huang2023high,grande2023topological}.

In the graph neural network literature, the relationship between pairs of points is quantified graphically.  
In the cases of latent trees, the relationships between points are induced by a weighted tree graph describing a simple relational structure present in the dataset.  The presents of an edge between any two points indicated a direct relationship between two nodes, and the weight of any such edge quantifying the strength of the relationship between the two connected, and thus related, nodes.  This relational structure can be interpreted as a latent hierarchy upon specifying a root node in the latent tree.  

\begin{figure}[H]%[hpt!]%
    \centering
    \begin{subfigure}[t]{.45\textwidth}
        \centering
        \includegraphics[width=1\linewidth]{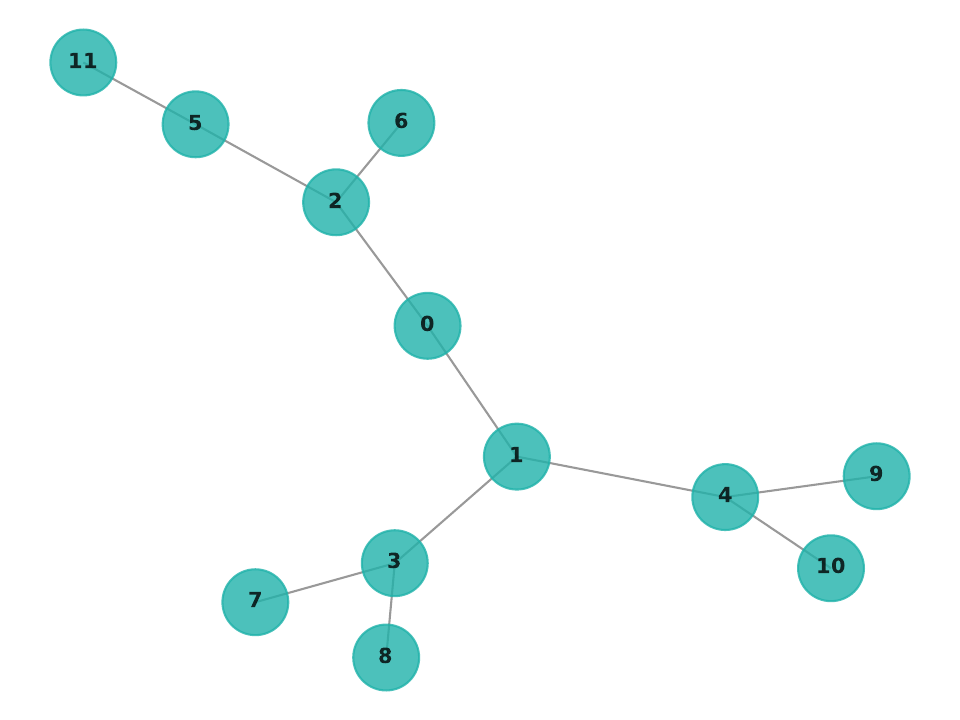}
        \caption{\label{fig:latent_tree_a}}  
    \end{subfigure}
    \centering
    \begin{subfigure}[t]{0.45\textwidth}
        \centering
        \includegraphics[width=.9\linewidth]{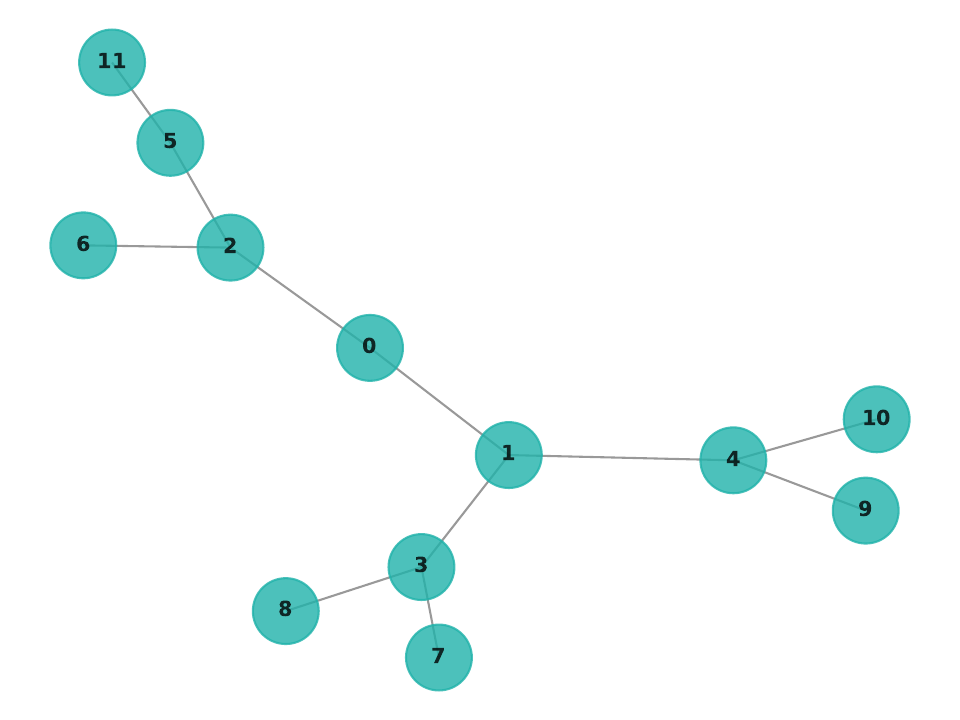}
        \caption{\label{fig:latent_tree_b}}
    \end{subfigure}
    \caption[1]{Figures~\ref{fig:latent_tree_a} and~\ref{fig:latent_tree_b} illustrate pointclouds in $\mathbb{R}^2$ with the same latent tree structure.  Both of these trees seem different when comparing their structure using the \textit{Euclidean distances}; however, instead, considering their latent tree structure reveals that they are identical as graphs.  This illustrates how Euclidean geometry often fails to detect the true latent (relational) geometry describing the hierarchical structure between points in a pointcloud.}
    \label{fig:latent_tree}
\end{figure}

Let $n$ be a positive integer and $V$ be a non-empty finite subset of $\mathbb{R}^n$, called a \textit{pointcloud}.  
Illustrated by Figure~\ref{fig:latent_tree}, a \textit{latent tree structure} on $V$ is a triple $(V,\mathcal{E},W)$ of a collection $\mathcal{E}$ of pairs $\{u,v\}$, called \textit{edges}, of $u,v\in V$ and an edge-weight map $\mathcal{W}:\mathcal{E}\rightarrow (0,\infty)$ satisfying the following property: For every distinct pair $u,v\in V$ there exists a unique sequence $u=u_0,\dots,u_i=v$ of distinct \textit{nodes} in $V$ such that the edges $\{u_0,u_1\},\dots,\{u_{i-1},u_{i}\}$ belong to $\mathcal{E}$; called a path from $u$ to $v$.  Thus, $\mathcal{T}=(V,\mathcal{E},\mathcal{W})$ is a \textit{finite weighted tree} with positive edge weights.  

Any latent tree structure $\mathcal{T}$ on $V$ induces a distance function, or metric, between the points of $V$.  This distance function, denoted by $d_{\mathcal{T}}$, measures the \textit{length} of the shortest path between pairs of points $u,v\in V$ and is defined by
\[
        d_{\mathcal{T}}(u,v)
    \eqdef 
        \inf
        \,
        \sum_{j=0}^{i-1}
        \,
            \mathcal{W}\big(\{u_j,u_{j+1}\}\big)
\]
where the infimum is taken over all sequences of paths $\{u_0,u_1\},\dots,\{u_{i-1},u_i\}$ from $u=u_0$ to $v=u_i$.  

If the weight function $\mathcal{W}(\{u,v\})=1$ for any edges $\{u,v\}\in \mathcal{E}$, then $\mathcal{T}$ is called a \textit{combinatorial tree}.  In which case, the distance between any two nodes $u,v\in V$ simplifies to the usual shortest path distance on an unweighted graph
\begin{equation}
\label{eq:definition_shortest_path_distance_on_unweighted_graph}
       d_{\mathcal{T}}(u,v)
    =
        \inf\big\{i\, : \,\exists\,\{v, v_1\},\dots\{v_{i-1},u\}\in \mathcal{E}\big\}
.
\end{equation}

The \textit{degree} of any point, or \textit{node/vertex}, $v\in V$ is the number of edges emanating from $v$; i.e.\ the cardinality of $\{\{u,w\}\in \mathcal{E}:\,v\in \{u,w\}\}$.  A node $v\in V$ is called a \textit{leaf} of the tree $\mathcal{T}$ if it has degree $1$.  E.g.\ in Figure~\ref{fig:tree}, all peripheral green points are leaves of the binary tree.

\subsection{Representations as Embeddings}
\label{s:Preliminaries__ss:Representations}
As in \cite{KratsiosDebarnot_2023_JMLR__UniversalEmbeddingsWasserstein}, a representation, or encoding, of latent tree structure on $V$ is simply a function $f: V\rightarrow \mathcal{R}$ into a space $\mathcal{R}$ equipped with a distance function $d_{\mathcal{R}}$, the pair $(\mathcal{R},d_{\mathcal{R}})$ of which is called a \textit{representation space}.  As in \cite{giovanni2022heterogeneous} representation $f$ is considered ``good'' if it accurately preserves the geometry of the latent tree structure $\mathcal{T}$ on $V$.  

Following the classical computer science literature, \cite{LinialLondonRavinovich_Combinatorica_1995__Embeddings,
Bartal_FOCS_1996__ProbApproxMetricEmbeddinsAlgoImplications,
RavinovichRaz_DiscCompGeom_1998__LowerBoundsDistortiionFiniteMetricSpaces,
AroraRaoVaziriani_JACM_2000__ExpandersGeomEmbeddingsGraphPartitioning,
Magen_RandApproxTechCS_2002__DimReduxVolPreerveAlgImplications,
STOC05Summary}, this means that $f$ is injective, or $1$-$1$, and its neither shrinks nor stretches the distances between pairs of nodes $u,v\in V$ when compared by $d_{\mathcal{R}}$.  For each $u,v\in V$ the following holds
\begin{equation}
\label{eq:epsilon_isometric_embedding}
        \alpha
        \,
        d_{\mathcal{T}}(u,v)
    \le 
        d_{\mathcal{R}}\big(
                f(u)
            ,
                f(v)
        \big)
    \le 
        \beta
        \,
        d_{\mathcal{T}}(u,v)
\end{equation}
where the constants $0< \alpha\le \beta<\infty$ are defined by
\[
     \beta
     \eqdef 
         \max_{ \substack{ u,v\in V \\ u\neq v} }\, \frac{d_{\mathcal{R}}\big(f(u),f(v)\big)}{d_{V}(u,v)},
    \mbox{ and }
    \alpha 
    \eqdef 
        \min_{ \substack{ u,v\in V \\ u\neq v } }\, \frac{d_{\mathcal{R}}\big(f(u),f(v)\big)}{d_{V}(u,v)}.
\]
These constants quantify the maximal shrinking ($\alpha$) and stretching ($\beta$) which $f$ exerts on the geometry induced by the latent tree structure $\mathcal{T}$ on $V$, and 
Note that, since $V$ is finite, then $0<\alpha\le \beta<\infty$ whenever $f$ is injective.

The total \textit{distortion} with which $f$ perturbs the tree structure $\mathcal{T}$ on $V$ is, denoted by $\operatorname{dist}(f)$, and defined by
\begin{equation}
\label{eq:distortion}
        \operatorname{dist}(f) 
    \eqdef 
        \begin{cases}
            \frac{\beta}{\alpha} & \mbox{ : if $f$ is injective} \\
            \infty & \mbox{ : otherwise}
        \end{cases}
\end{equation}
We say that a tree $\mathcal{T}$ can be \textit{asymptotically isometrically represented} in $\mathcal{R}$ if there is a sequence $(f_n)_{n=1}^{\infty}$ of maps from $V$ to $\mathcal{R}$ whose distortion is asymptotically optimal distortion; i.e.\ $\lim\limits_{n\rightarrow \infty}\, \operatorname{dist}(f_n)=1$. We note that a sequence of embeddings $f_n$ need not have a limiting function mapping $V$ to $\mathcal{R}$ even if its distortion converges to $1$; in particular, $(f_n)_{n\in \mathbb{N}}$ need not converge to an isometry.

\section{Graph Representation Learning Results}
\label{s:Main_Result}

This section contains our main result, which establishes the main motivation behind HNNs, namely the belief that they represent trees to arbitrary precision in a two dimensional hyperbolic space.

\subsection{Lower-Bounds on Distortion for MLP Embeddings of Latent Trees}
\label{s:LowerBounds}
The power of HNNs is best appreciated when juxtaposed against the \textit{lower} bounds minimal distortion implementable by any MLP embedding any large tree into a low-dimensional Euclidean space.  In particular, there cannot exist any MLP model, regardless of its depth, width, or (possibly discontinuous) choice of activation function, which can outperform the embedding of a sufficiently overparameterized HNN using only a two-dimensional representation space.  

\begin{thm}[Lower-Bounds on the Distortion of Trees Embedded by MLPs]
\label{thm:LowerBound}
Let $L,n,d\in \mathbb{N}_+$, and fix an activation function $\sigma:\mathbb{R}\rightarrow\mathbb{R}$.  
For any finite $V\subset \mathbb{R}^n$ with a latent combinatorial tree structure $\mathcal{T}=(V,\mathcal{E},\mathcal{W})$ having $L>2^d$ leaves and if $f:\mathbb{R}^n\rightarrow \mathbb{R}^d$ is an MLP with activation function $\sigma$, satisfying
\[
        \alpha
        \,
        d_{\mathcal{T}}(u,v)
    \le 
        \|
            f(u)
            -
            f(v)
        \|
    \le 
        \beta
        \,
        d_{\mathcal{T}}(u,v)
\]
for all $u,v\in V$ and some $0< \alpha\le \beta$ independent of $u$ and $v$ then, $f$
incurs a distortion $\operatorname{dist}(f)$ of at least
\[
        \operatorname{dist}(f)
    \ge 
        \Omega(L^{1/d})
    .
\]
The constant suppressed by $\Omega$, is independent of the depth, width, number of trainable parameters, and the activation function $\sigma$ defining the MLP.
\end{thm}

Theorem~\ref{thm:LowerBound} implies that if $V\subseteq \mathbb{R}^n$ is large enough and has a latent tree structure with $\Omega(4^{d^2})$ leaves then any MLP $f:\mathbb{R}^n\rightarrow\mathbb{R}^d$ cannot represent $(V,d_{\mathcal{T}})$ with a distortion of less than $\Omega(4^d)$.  Therefore, if $d$, $\#V$, and $L$ are large enough, any MLP must represent the latent tree structure on $V$ arbitrarily poorly.  
We point out that their MLP's structure alone is not the cause of this limitation since we have not imposed any structural constraints on its depth, width, number of trainable parameters, or its activation function; instead, the incompatibility between the geometry of a tree and that of a Euclidean space, which no MLP can resolve.

\subsection{Upper-Bounds on the Complexity of HNNs Embeddings of Latent Trees}
\label{s:UpperBounds}

Our main positive result shows that the HNN model~\ref{defn:Hyperbolic_NNs} can represent any pointcloud with a latent tree structure into the hyperbolic space $\mathbb{H}_{\kappa}^d$ with an arbitrarily small distortion by a low-capacity HNN.

\begin{thm}[HNNs Can Asymptotically Isometrically Represent Latent Trees]
\label{thm:Main}
Fix $n,d,N\in \mathbb{N}_+$ with $d\ge 2$ and fix $\lambda > 1$.  For any $N$-point subset $V$ of $\mathbb{R}^n$ and any latent tree structure $\mathcal{T}=(V,\mathcal{E},\mathcal{W})$ on $V$ of degree at least $2$, there exists a curvature parameter $\kappa<0$ and an HNN $f:\mathbb{R}^n\rightarrow \mathbb{H}^d_{\kappa}$ such that
\[
    \frac1{\lambda}
        \,
        d_{\mathcal{T}}(u,v)
    \le
        d_{\kappa}(f(u),f(v))
    \le
        \lambda
        \,
        d_{\mathcal{T}}(u,v)
\]
holds for each pair of $u,v\in V$.  Moreover, the depth, width, and number of trainable parameters defining $f$ are independent of $\lambda$; recorded in Table~\ref{tab:Complexities_HNN}.  
\end{thm}

Theorem~\ref{thm:Main} considers HNNs with the typical $\operatorname{ReLU}$ activation function.  However, using an argument as in \citep[Proposition 1]{yarotsky2017error} the result can likely be extended to any other continuous piece-wise linear activation function with at least one piece/break, e.g\ $\operatorname{PReLU}$.  Just as in \citep[Proposition 1]{yarotsky2017error}, such modifications should only scale the network depth, width, and the number of its trainable parameters up by a constant factor depending only on the number of pieces of the chosen piece-wise linear activation function.

Since the size of the tree in Theorem~\ref{thm:Main} did not constrain the embedding quality of an HNN, we immediately deduce the following corollary which we juxtapose against Theorem~\ref{thm:LowerBound}.
\begin{cor}[HNNs Can Asymptotically Embed Large Trees]
\label{cor:LowerBound__Comparable}
Let $L,n,d\in \mathbb{N}_+$ with $d\ge 2$.
For any finite $V\subset \mathbb{R}^n$ with a latent combinatorial tree structure $\mathcal{T}=(V,\mathcal{E},\mathcal{W})$ with $L>2^d$ leaves, and any $r>0$, there exists a curvature parameter $\kappa<0$ and an HNN $f:\mathbb{R}^n\rightarrow\mathbb{H}^d_{\kappa}$ satisfying
\[
    \operatorname{dist}(f) \le \frac1{L^r}
.
\]
\end{cor}

\begin{table}[H]%[hbt!]%
    \centering
    \caption{Estimates on the Complexity of the HNN implementing the $N$-point memorization of the function $f^{\star}:\mathbb{R}^n\rightarrow \mathbb{H}^d$ in Lemma~\ref{lem:MemorizationInterpolation}.}
        \resizebox{\columnwidth}{!}{%
        \renewcommand{\arraystretch}{1.3}
    		\begin{tabular}{@{}ll@{}}
    			\cmidrule[0.3ex](){1-2}
    			\textbf{Complexity} & \textbf{Upper-Bound} 
    			\\
    			\midrule
        		      Depth 
                    & 
                        $
                             \mathcal{O}\left( 
                                N \left\{ 
                                    1+\sqrt{N\log{N}} \left[ 
                                        1+\frac{\log(2)}{\log(n)} \left(
                                            C_n+\frac{\log\left( N^2 \operatorname{aspect} (V,\|\cdot\|_2 \right)}{\log(2)}
                                        \right)_+
                                    \right]
                                \right\} 
                            \right)
                        $
                \\
                        Width
                    & 
                        $
                            n(N-1)+\max\{d,12\}
                        $
                \\
                        % \multirow{2}{*}{
                        N. Train. Par.
                        % }
                    & 
                        $
                        \mathcal{O}\left(
                                N\left( \frac{11}{4}\max\{n,d\}N^2 - 1\right) 
                                \left\{
                                    d+\sqrt{N\log{N}}\left[ 
                                        1+\frac{\log(2)}{\log(n)} 
                            \right.\right.\right.
                            $
                \\
                &
                            $
                            \left.\left.
                                    \times \left(
                                        C_n+\frac{\log\left( N^2 \operatorname{aspect} (V,\|\cdot\|_2 \right)}{\log(2)}
                                    \right)_+
                                \right] \max\{d,12\} (1+\max\{d,12\})
                            \right\}
                            $
                \\
                &   
                            $
                            +
                            \left.
                            N \left\{ 
                                1+\sqrt{N\log{N}} \left[ 
                                    1+\frac{\log(2)}{\log(n)} \left(
                                        C_n+\frac{\log\left( N^2 \operatorname{aspect} (V,\|\cdot\|_2 \right)}{\log(2)}
                                    \right)_+
                                \right]
                            \right\} 
                        \right)
                        $
                \\
    			\bottomrule
    		\end{tabular}
    		}% END Resize box
        \label{tab:Complexities_HNN}
        \caption*{The “dimensional constant” is $0<C_n=\frac{2\log (5\sqrt{2\pi}) + \frac{3}{2}\log n - \frac{1}{2}\log (n+1)}{2\log(2)}$.}
    \end{table}

\subsection{Experimental Illustrations}
\label{s:Experiments}

To gauge the validity of our theoretical results, we conduct a performance analysis for tree embedding. We compare the performance of HNNs with that of MLPs through a sequence of synthetic graph embedding experiments. Our primary focus lies on binary, ternary, and random trees. For the sake of an equitable comparison, we contrast MLPs and HNNs employing an equal number of parameters. Specifically, all models incorporate 10 blocks of linear layers, accompanied by batch normalization and ReLU activations, featuring 100 nodes in each hidden layer. The training process spans 10 epochs for all models, employing a batch size of 100,000 and a learning rate of $10^{-2}$. The $x$ and $y$ coordinates of the graph nodes in $\mathbb{R}^{2}$ are fed into both the MLP and HNN networks, which are tasked to map them to a new embedding space. An algorithm is used to generate input coordinates, simulating a force-directed layout of the tree. In this simulation, edges are treated as springs, pulling nodes together, while nodes are treated as objects with repelling forces akin to an anti-gravity effect. This simulation iterates until the positions reach a state of equilibrium. The algorithm can be reproduced using the \texttt{NetworkX} library and the \texttt{spring layout} for the graph.

Counting the neighbourhood hops as in \ref{eq:definition_shortest_path_distance_on_unweighted_graph} defines the distance between nodes, resulting in a scalar value. The networks must discover a suitable representation to estimate this distance. We update the networks based on the MSE loss comparing the actual distance between nodes, $d_{true}$, to the predicted distance based on the network mappings, $d_{pred}$:
\begin{equation}
    Loss = MSE(d_{true},d_{pred}).
\end{equation}
In the case of the MLP the predicted distance is computed using:
\begin{equation}
\label{eq:MLP_Loss}
    d_{pred} = \|MLP(x_1,y_1)-MLP(x_2,y_2)\|_2,
\end{equation}
where $(x_1,y_1)$ and $(x_2,y_2)$ are the coordinates in $\mathbb{R}^2$ of a synthetically generated latent tree (which may be binary, ternary or random). For the HNN we use the loss function
\begin{equation}
\label{eq:HNN_Loss}
    d_{pred} = d_{-1}(HNN(x_1,y_1),HNN(x_2,y_2)).
\end{equation}
In the case of the HNN, we use the hyperboloid model, with an exponential map at the pole, to map the representations to hyperbolic space. In particular, we do not even require any hyperbolic biases to observe the gap in performance between the MLP and HNN models, which are trained to embed the latent trees by respectively optimizing the loss functions~\ref{eq:MLP_Loss} and~\ref{eq:HNN_Loss}.

\begin{figure}[htbp]
    \centering
    \begin{subfigure}[b]{0.32\textwidth}
        \centering
        \includegraphics[width=\textwidth]{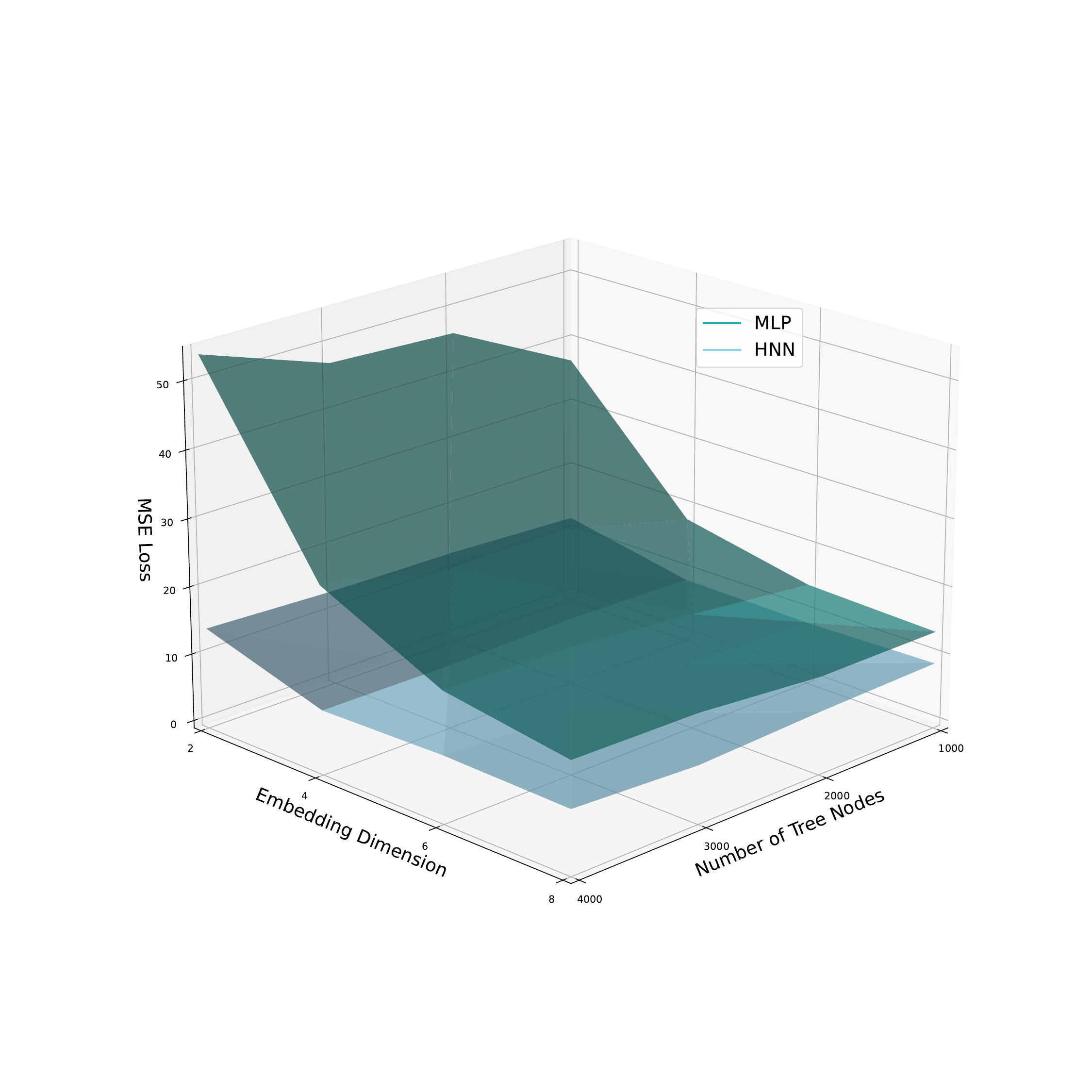}
        \caption{Binary Tree}
        \label{fig:binarytree_loss_surface}
    \end{subfigure}
    \hfill
    \begin{subfigure}[b]{0.32\textwidth}
        \centering
        \includegraphics[width=\textwidth]{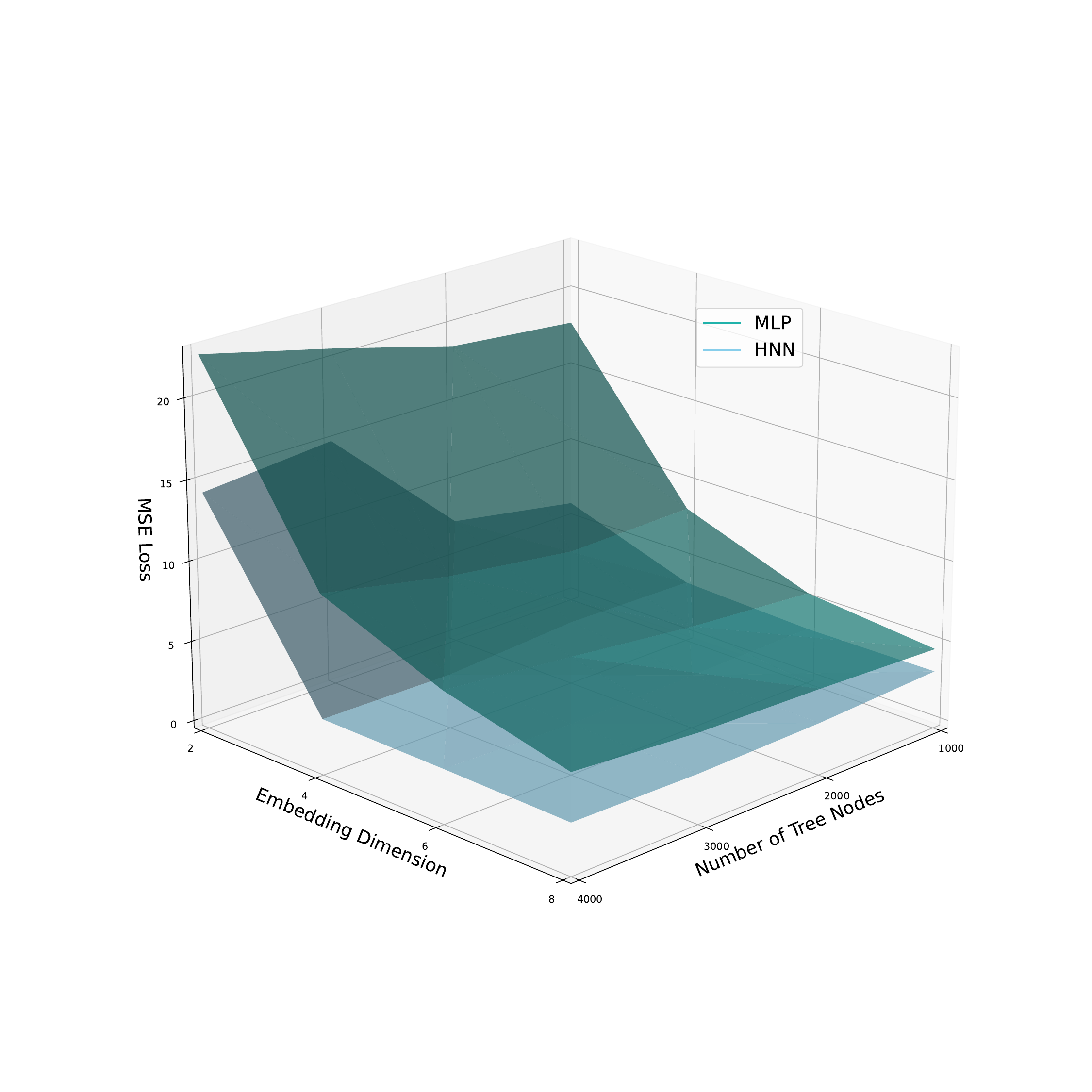}
        \caption{Ternary Tree}
        \label{fig:ternary_loss_surface}
    \end{subfigure}
    \hfill
    \begin{subfigure}[b]{0.32\textwidth}
        \centering
        \includegraphics[width=\textwidth]{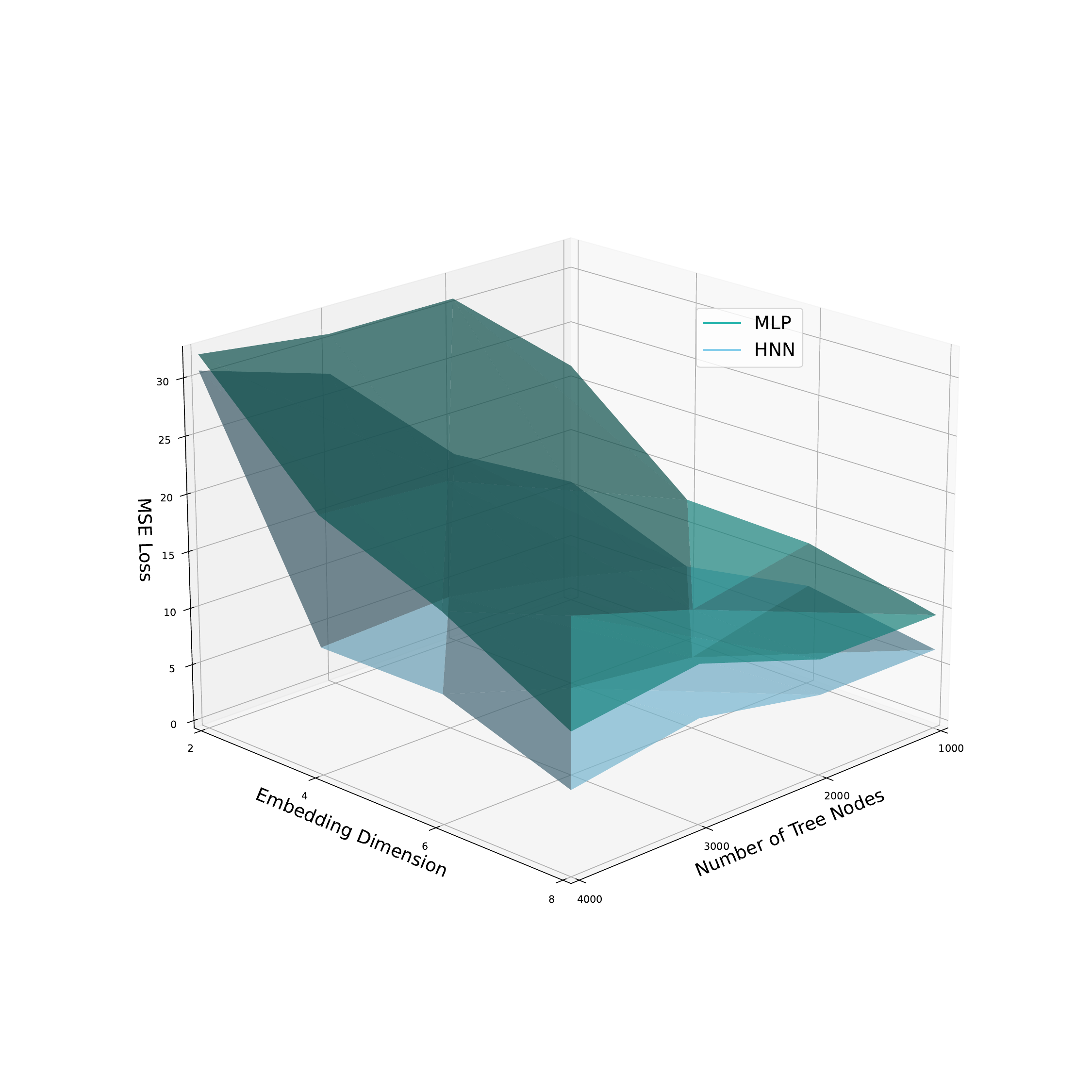}
        \caption{Random Tree}
        \label{fig:randomtree_loss_surface}
    \end{subfigure}
    \caption{Tree embedding error surfaces for different trees using MLPs and HNNs.}
    \label{fig:loss_surfaces}
\end{figure}

We conduct embedding experiments on graphs ranging from 1,000 to 4,000 nodes, and we assess the impact of employing various dimensions for the tree embedding spaces. Specifically, we explore dimensionalities multiples of 2, ranging from 2 to 8. In Figure \ref{fig:loss_surfaces}, we can observe that HNNs consistently outperform MLPs at embedding trees, achieving a lower MSE error in all configurations.

We now overview the derivation of our upper bounds on the embedding capabilities HNNs with $\operatorname{ReLU}$ activation function and our lower-bounds on MLPs with any activation function for pointclouds with latent tree structure.

\section{Theoretical analysis}
\label{s:Proof}
We first prove Theorem~\ref{thm:LowerBound} which follows relatively quickly from the classical results of \cite{Matouvek_1999_IJM__EmbeddingsTreesUCBanSpaces} and \cite{Gupta_2000_ACM__QuantitiativefinitedimembeddingsTrees} from the metric embedding theory and computer science literature.
We then outline the proof of Theorem~\ref{thm:Main}, which is much more involved, both technically and geometrically, the details of which are relegated to Section~\ref{s:Proof}.  
Corollary~\ref{cor:LowerBound__Comparable} is directly deduced from Theorem~\ref{thm:Main}.

\subsection{Proof of the Lower-Bound\texorpdfstring{ - {Theorem~\ref{thm:LowerBound}}}{}}
We begin by discussing the proof of the lower-bound.
\begin{proof}[{Proof of Theorem~\ref{thm:LowerBound}}]
Any MLP $f:\mathbb{R}^n\rightarrow\mathbb{R}^d$ with any activation function $\sigma:\mathbb{R}\rightarrow\mathbb{R}$ for which there exists constants $0<\alpha\le \beta$ satisfying
\[
        \alpha
        \,
        d_{\mathcal{T}}(u,v)
    \le 
        \|
            f(u)
            -
            f(v)
        \|
    \le 
        \beta
        \,
        d_{\mathcal{T}}(u,v)
\]
defines a bi-Lipschitz embedding of the tree-metric space $(V,d_{\mathcal{T}})$ into the $d$-dimensional Euclidean space.  

Since $L>2^d$ then we may apply \citep[Proposition 5.1]{Gupta_2000_ACM__QuantitiativefinitedimembeddingsTrees}, which is a version of a negative result of  \cite{Bourgain_1986_IJM__EmbeddingMetricSpacesSuperreflexivBanachSpaces} and of \cite{Matouvek_1999_IJM__EmbeddingsTreesUCBanSpaces} for non-embeddability of large trees in Euclidean space for general trees.  The result, namely \citep[Proposition 5.1]{Gupta_2000_ACM__QuantitiativefinitedimembeddingsTrees}, implies that any bi-Lipschitz embedding of $(V,d_{\mathcal{T}})$ into $(\mathbb{R}^d,\|\cdot\|_2)$ must incur a distortion no less than $\Omega(L^{1/d})$; in particular, this is the case for $f$.  Therefore, $\frac{\beta}{\alpha} \ge \Omega(L^{1/d})$.
\end{proof}
The proof of the lower-bound shows that there cannot be \textit{any} deep learning model which injectively maps $(V,d_{\mathcal{T}})$ into a $d$-dimensional Euclidean space with a distortion of less than $\Omega(L^{1/d})$.  

\subsection{{Proof of \texorpdfstring{Theorem~\ref{thm:Main}}{Main Result}}}
\label{s:Proof__ss:UpperBound}

We showcase the three critical steps in deriving theorem~\ref{thm:Main}.  First, we show that HNNs can implement, i.e.\ memorize, arbitrary functions from any $\mathbb{R}^n$ to any $\mathbb{H}^d$, for arbitrary integer dimensions $n$ and $d$.  Second, we construct a sequence of embeddings with, whose distortion asymptotically tends to $1$, in a sequence of hyperbolic spaces $(\mathbb{H}^d_{\kappa},d_{\kappa})$ of arbitrarily large sectional curvature $\kappa$.  We then apply our memorization result to deduce that these ``asymptotically isometric'' embeddings can be implemented by HNNs, quantitatively.  

The proofs of both main lemmata are relegated to Section~\ref{s:Proof__ss:Details}.  

\paragraph{Step 1 - Exhibiting a Memorizing HNN and Estimating its Capacity}
We first need the following quantitative memorization guarantee for HNNs.

\begin{lem}[Upper-Bound on the Memory Capacity of an HNN]
\label{lem:MemorizationInterpolation}
Fix $N,n,d\in \mathbb{N}_+$.
For any $N$-point subset $V\subset \mathbb{R}^n$ and any function $f^{\star}:\mathbb{R}^n \rightarrow \mathbb{H}^d$, there exists an HNN $f:\mathbb{R}^n \rightarrow \mathbb{H}^d$ satisfying 
\[
        f(v)
    =
        f^{\star}(v)
\]
for each $v\in V$.  Moreover, the depth, width, and number of trainable parameters defining $f^{\star}$ are bounded-above in Table~\ref{tab:Complexities_HNN}.  
\end{lem}

The capacity estimates for the HNNs constructed in Theorem~\ref{thm:Main} and its supporting Lemma~\ref{lem:MemorizationInterpolation} depend on the configuration of the pointcloud $V$ in $\mathbb{R}^n$ with respect to the Euclidean geometry of $\mathbb{R}^n$.  The configuration is quantified by the ratio of the largest distance between distinct points over the smallest distance between distinct points, called the \emph{aspect ratio} on \citep[page 9]{KratsiosDebarnot_2023_JMLR__UniversalEmbeddingsWasserstein}, also called the separation condition in the MLP memorization literature; e.g.~\citep[Definition 1]{Sublinear_Memorization}.  
\[
        \operatorname{aspect}(V)
    \eqdef
        \frac{\max_{x,\Tilde{x}\in V} \|x-\Tilde{x}\|_2}{\min_{x,\Tilde{x}\in V ; x\ne\Tilde{x}} \|x-\Tilde{x}\|_2}
    .
\]
Variants of the aspect ratio have also appeared in computer science, e.g.~\citep{Various_WorkshopProceedings_2001__ApproximationRandomizationCombinatorialOptimization,newman2023online} and the related metric embedding literature; e.g.~\citep{KrauthgameLeeNaor2004}.

\paragraph{Step 2 - Constructing An Asymptotically Optimal Embedding Into \texorpdfstring{$\mathbb{H}^2_{\kappa}$}{The Hyperbolic Plane With Sufficiently Negative Curvature}}

\begin{lem}[HNNs Universally Embed Trees Into Hyperbolic Spaces]
\label{lem:ConstructionEmbeddingAtInfinity}
Fix $n,d,N\in \mathbb{N}_+$ with $d\ge 2$ and fix $\lambda > 1$.  For any $N$-point subset $V$ of $\mathbb{R}^n$ and any latent tree structure $\mathcal{T}=(V,\mathcal{E},\mathcal{W})$ on $V$ of degree at least $2$, there exists a map $f^{\star}:\mathbb{R}^n\rightarrow \mathbb{H}^d$ and a sectional curvature $\kappa<0$ satisfying
\begin{equation}
\label{eq:lem:ConstructionEmbeddingAtInfinity_Bound}
        \frac1{\lambda}
        \,
        d_{\mathcal{T}}(u,v)
    <
        d_{\kappa}(f^{\star}(u),f^{\star}(v))
    <
        \lambda
        \,
        d_{\mathcal{T}}(u,v)
\end{equation}
for each $u,v\in V$.  Furthermore, $\kappa$ tends to $-\infty$ as $\lambda$ tends to $1$.
\end{lem}

\paragraph{Step 3 - Memorizing the Embedding Into \texorpdfstring{$\mathbb{H}^d_{\kappa}$}{The Hyperbolic Plane With Sufficiently Negative Curvature} with an HNN}

\begin{proof}[{Proof of Theorem~\ref{thm:Main}}]
Fix $n,d,N\in \mathbb{N}_+$ with $d\ge 2$ and $\lambda >1$.  Let $V$ be an $N$-point subset of $\mathbb{R}^n$ and $\mathcal{T}=(V,\mathcal{E},\mathcal{W})$ be a latent tree structure on $V$ of degree at least $2$.  By Lemma~\ref{lem:ConstructionEmbeddingAtInfinity}, there exists a $\kappa<0$ and a $\lambda$-isometric embedding $f^{\star}:(V,d_{\mathcal{T}})\rightarrow (\mathbb{H}_{\kappa}^d,d_{\kappa})$; i.e.~\eqref{eq:lem:ConstructionEmbeddingAtInfinity_Bound} holds.  

Since $V$ is a non-empty finite subset of $\mathbb{R}^n$, we may apply Lemma~\ref{lem:MemorizationInterpolation} to infer that there exists an HNN $f:\mathbb{R}^n\rightarrow\mathbb{H}^d$ satisfying $f(v)=f^{\star}(v)$ for each $v\in V$.  Furthermore, its depth, width, and the number of its trainable parameters is recorded in Table~\ref{tab:Complexities_HNN}.  This conclude our proof.
\end{proof}

\begin{proof}[{Proof of Corollary~\ref{cor:LowerBound__Comparable}}]
The the result follows upon taking $\lambda = L^{-r/2}$ in Theorem~\ref{thm:Main}.
\end{proof}

\subsection{Details on the Proof of the Upper-Bound\texorpdfstring{ In Theorem~\ref{thm:Main}}{}}
\label{s:Proof__ss:Details}

We now provide the explicit derivations of all the above lemmata used to prove Theorem~\ref{thm:Main}.

\begin{proof}[{Proof of Lemma~\ref{lem:MemorizationInterpolation}}]
\hfill\\
\textbf{Overview:}
\textit{The proof of this lemma can be broken down into $4$ steps.  
First, we linearized the function $f^{\star}$ to be memorized by associating it to a function between Euclidean spaces.  
Next we memorize the transformed function in Euclidean using a MLP with $\operatorname{ReLU}$ activation function which we then transform to an $\mathbb{H}^d$-valued function which memorizes $f^{\star}$.  
We then show that this transformed MLP can be implemented by an HNN.  
Finally, we tally the parameters of this HNN representation of the transformed MLP.}
\hfill\\
\noindent\textit{Step 1 - Standardizing Inputs and Outputs of the Function to be Memorized}
\hfill\\
Fix any $y\in \mathbb{H}^d$.  Since $\mathbb{H}_{\kappa}^d$ is a simply connected Riemannian manifold of non-positive curvature then the Cartan-Hadamard Theorem, as formulated in \citep[Corollary 6.9.1]{Jost_2017_Book__RiemannianGeometryGeometricAnalysis}, implies that the map $\Exp{x}:T_x(\mathbb{H}_{\kappa}^d)\rightarrow \mathbb{H}_{\kappa}^d$ is global diffeomorphism.  Therefore, the map $\Log{x}: \mathbb{H}_{\kappa}^d \rightarrow T_x(\mathbb{H}_{\kappa}^d)$ is well-defined and a bijection.  In particular, this is the case for $x={\boldsymbol{1}_d}$.  
Therefore, the map $\pi_d\circ \Log{{\boldsymbol{1}_d}}:\mathbb{H}^d\rightarrow \mathbb{R}^d$ is a bijection.  Consider the map $\bar{f}: \mathbb{R}^n\rightarrow \mathbb{R}^d$ defined by
\[
    \bar{f}\eqdef \pi_d\circ  \Log{{\boldsymbol{1}_d}} \circ f^{\star}
    .
\]
Note that, since $\pi_d:T_{{\boldsymbol{1}_d}}(\mathbb{H}^d)\rightarrow\mathbb{R}^d$ is a linear isomorphism it is a bijection and $\iota_d$ is its two-sided inverse.  Therefore, the definition of $\bar{f}$ implies that
\begin{equation}
\label{PRF_lem:Interpolation__eq:InvertingfStarAndMappingBack}
        (\Exp{1_d}\circ \iota_d)
    \circ 
        \bar{f}
    =
        f^{\star}
.
\end{equation}
\noindent\textit{Step 2 - Memorizing the Standardized Function}
\hfill\\
    Since $V\subseteq \mathbb{R}^n$ and $\bar{f}:\mathbb{R}^n\rightarrow \mathbb{R}^d$ then we may apply \citep[Lemma 20]{KratsiosDebarnot_2023_JMLR__UniversalEmbeddingsWasserstein} to deduce that there is an MLP (feedforward neural network) with $\operatorname{ReLU}$ activation function $\tilde{f}:\mathbb{R}^n\rightarrow \mathbb{R}^d$ that interpolates $\bar{f}$; i.e.~there are positive integers $I,n=d_0,\dots,d_{I+2}=d\in \mathbb{N}_+$, such that for each $i=1,\dots,I+1$ there is a $d_{i+1}\times d_i$ matrix $A^{(i)}$ and a vector $b^{(i)}\in \mathbb{R}^{d_{i+1}}$ implementing the representation
    \begin{equation}
    \label{PRF_lem:Interpolation__eq:ReLU_Definition}
    \begin{aligned}
        % Readout/Final Layer
            \tilde{f}(u)
        &
        =
                A^{(I+1)}
                \,
                u^{(I)}
            +
                b^{(I+1)}
    \\
        % Hidden Layers
            u^{(i)}
        & 
            \operatorname{ReLU}\bullet
            (
                    A^{(i)}u^{(i-1)}
                +
                    b^{(i)}
            )
    \\
        %% Feature Representation
            u^{(0)} 
        & 
        =
            u
    \end{aligned}
    \end{equation}
    for each $v\in \mathbb{R}^n$, and satisfying the interpolation/memorization condition
    \begin{equation}
    \label{PRF_lem:Interpolation__eq:MemorizedByReLU}
            \tilde{f}(v)
        =
            \bar{f}(v)
    \end{equation}
    for each $v\in V$.  Furthermore, its depth, width, and number of non-zero/trainable parameters are 
    \begin{enumerate}
        \item The \textit{width} of $\tilde{f}$ is $n(N-1)+\max\{d,12\}$,
        \item The \textit{depth} ($I$) of $\tilde{f}$ is 
        \[
            \mathcal{O}\left( 
                N \left\{ 
                    1+\sqrt{N\log{N}} \left[ 
                        1+\frac{\log(2)}{\log(n)} \left(
                            C_n+\frac{\log\left( N^2 \operatorname{aspect} (V,\|\cdot\|_2 \right)}{\log(2)}
                        \right)_+
                    \right]
                \right\} 
            \right),
        \]
        \item The \textit{number of (non-zero) trainable parameters} of $\tilde{f}$ is 
        \[
        \begin{aligned}
                & \mathcal{O}\left(
                    N\left( \frac{11}{4}\max\{n,d\}N^2 - 1\right) 
                    \left\{
                        d+\sqrt{N\log{N}}\left[ 
                            1+\frac{\log(2)}{\log(n)} 
            \right.\right.\right.
            \\
            &
            \left.\left.\left.
                            \times \left(
                                C_n+\frac{\log\left( N^2 \operatorname{aspect} (V,\|\cdot\|_2 \right)}{\log(2)}
                            \right)_+
                        \right] \max\{d,12\} (1+\max\{d,12\})
                    \right\}
                \right)
        .
        \end{aligned}
        \]
    \end{enumerate}
    \textbf{Comment:} \textit{In the proof of the main result, the aspect ratio $\operatorname{aspect}(V)$ will not considered with respect to the shortest path distance $d_{\mathcal{T}}$ on $V$, given by its latent tree structure, but rather with respect to the Euclidean distance $\|\cdot\|_2$ on $\mathbb{R}^n$.  This is because, the only role of the MLP $\tilde{f}$ is to interpolate pairs of points of the function $\bar{f}$ denoted between Euclidean spaces.  The ability of an MLP to do so depends on how close those points are to one another in the Euclidean sense.}

    Combining~\eqref{PRF_lem:Interpolation__eq:InvertingfStarAndMappingBack} with~\eqref{PRF_lem:Interpolation__eq:MemorizedByReLU}, with the fact that $\Exp{x}\circ \iota_d$ is a bijection implies that the following
    \begin{equation}
    \label{PRF_lem:Interpolation__eq:MemorizedByPreRepresentedHNN}
    \begin{aligned}
                f^{\star}(v)
        = & 
                (\Exp{1_d}\circ \iota_d)
            \circ 
                \bar{f}(v)
    \\
        = &
                (\Exp{1_d}\circ \iota_d)
            \circ 
                \tilde{f}(v)
    \\
        = & 
                (\Exp{1_d}\circ \iota_d)
            \circ 
                (
                    \tilde{f}
                \circ 
                    \Log{{\boldsymbol{1_{n}}}}
                )
            \circ 
                \Exp{\boldsymbol{1_{n}}}
                (v)    
    \end{aligned}
    \end{equation}
    holds for every $v\in V$.  
    It remains to be shown that the function on the right-hand side of~\eqref{PRF_lem:Interpolation__eq:MemorizedByPreRepresentedHNN} can be implemented by an HNN.

% \hfill\\
\noindent\textit{Step 3 - Representing $ \Exp{1_d}\circ \iota_d\circ \tilde{f}(x)$ as an HNN}
\hfill\\
For $i=0,\dots,I+1$ set $c^{(i)}\eqdef {\boldsymbol{1}_{d_i}}$.  Observe that, for each $i=1,\dots,I$, $\Exp{c^{(i)}}\circ \Log{c^{(i)}} = 1_{T_{c^{(i)}}(\mathbb{H}^{d_i})}$ and that the following holds
\begin{align}
\notag
        \underline{\oplus}_{c^{(i)}}\circ \overline{\oplus}^{c^{(i)}}
    = & 
            \pi_{d^{(i)}}
        \circ
            P_{c^{(i)} \mapsto  {\boldsymbol{1}_{d_i}} }
        \circ 
            P_{ {\boldsymbol{1}_{d_i}} \mapsto c^{(i)} }
        \circ 
            \iota_{d^{(i)}}
\\
\label{PRF_lem:Interpolation__eq:cito1di}
    = & 
            \pi_{d^{(i)}}
        \circ
            P_{{\boldsymbol{1}_{d_i}} \mapsto  {\boldsymbol{1}_{d_i}} }
        \circ 
            P_{ {\boldsymbol{1}_{d_i}} \mapsto {\boldsymbol{1}_{d_i}}}
        \circ 
            \iota_{d^{(i)}}
\\
\label{PRF_lem:Interpolation__eq:cito1di_itendity}
    = & 
            \pi_{d^{(i)}}
        \circ
            1_{T_{\boldsymbol{1}_{d_i}}(\mathbb{H}^{d_i})}
        \circ 
            1_{T_{\boldsymbol{1}_{d_i}}(\mathbb{H}^{d_i})}
        \circ 
            \iota_{d^{(i)}}
\\
\notag
    = & 
            \pi_{d^{(i)}}
        \circ 
            \iota_{d^{(i)}}
\\
\notag
    = & 
        1_{\mathbb{R}^{d_i}}
.
\end{align}
where~\eqref{PRF_lem:Interpolation__eq:cito1di} follows from our definition of $c^{(i)}$ and~\eqref{PRF_lem:Interpolation__eq:cito1di_itendity} follows since parallel transport from $T_{{\boldsymbol{1}_{d_i}}}(\mathbb{H}^{d_i})$ to itself along the unique distance minimizing curve (geodesic) $\gamma:[0,1]\rightarrow \mathbb{H}^{d_i}$ emanating from and terminating at ${\boldsymbol{1}_{d_i}}$; namely, $\gamma(t)={\boldsymbol{1}_{d_i}}$ for all $0\le t\le 1$.  Therefore, any HNN with representation of an HNN in Definition~\ref{defn:Hyperbolic_NNs}, with these specifications of $c^{(0)}$, $\dots$, $c^{(I+1)}$ can be represented as
\[
                (\Exp{1_d}\circ \iota_d)
            \circ 
                (
                    g
                \circ 
                    \Log{{\boldsymbol{1_{n}}}}
                )
            \circ 
                \Exp{\boldsymbol{1_{n}}}
\]
where the map $g:\mathbb{R}^n\rightarrow\mathbb{R}^d$ is an MLP with ReLU activation function; i.e.\ it can be represented as
    \begin{equation}
    \label{PRF_lem:Interpolation__eq:ReLU_Definition___ReducedRepresentation}
    \begin{aligned}
        % Readout/Final Layer
            g(u)
        &
        =
            \tilde{A}^{(I+1)}
            \,
            (
                u^{(I)}
            +
                \tilde{b}^{(\tilde{I}+1)}
            )
    \\
        % Hidden Layers
            u^{(i)}
        & 
            \operatorname{ReLU}\bullet
            \big(
                    \tilde{A}^{(i)}
                +
                    \tilde{b}^{(i)}
            \big)
    \\
        %% Feature Representation
            u^{(0)} 
        & 
        =
            u
    \end{aligned}
    \end{equation}
for some integers $\tilde{I},n=d_0,\dots,d_{I+2}=d\in \mathbb{N}_+$, $\tilde{d}_{i+1}\times \tilde{d}_i$ matrices $\tilde{A}^{(i)}$ and a vectors $\tilde{b}^{(i)}\in \mathbb{R}^{\tilde{d}_{i+1}}$.  Setting $g\eqdef \tilde{f}$, implies that the map $
        (\Exp{x}\circ \iota_d)
    \circ 
        (
            \tilde{f}
        \circ 
            \Exp{\boldsymbol{1_{n}}}
        )
    \circ 
        \Log{{\boldsymbol{1_{n}}}}
$ in~\eqref{PRF_lem:Interpolation__eq:MemorizedByPreRepresentedHNN} defines an HNN.

\noindent\textit{Step 4 - Tallying Trainable Parameters}
\hfill\\
By construction the depth and width of $f$ respectively equal to the depth and width of $\tilde{f}$.  The number of parameters defining $f$ equal to the number of parameters defining $\tilde{f}$ plus $I+1$, since   
$\|c^{(i)}\|_0=1$ for each $i=0,\dots,I+1$.
\end{proof}

Our proof of Lemma~\ref{lem:ConstructionEmbeddingAtInfinity} relies on some concepts from metric geometry, which we now gather here, before deriving the result.  

A Geometric realization of a (positively) weighted graph $G$ can be seen as the metric space obtained by gluing together real intervals of lengths equal to corresponding weights at corresponding endpoints, according to the pattern of $G$, with the shortest path distance. Following \citep[Definition 3.1.1]{DasSimmonsUrbanski_2017_AMSBook_GeometryDynamicsGromovHyperbolicSpaces}, this is formalized as the following metric space.

\begin{definition}[Geometric Realization Of A Weighted Tree] 
A geometric realization of a weighted tree $\mathcal{T}=(V,\mathcal{E},\mathcal{W})$ is the metric space $(X_{\mathcal{T}},d_{X_{\mathcal{T}}})$ whose pointset 
$X_{\mathcal{T}}=V\cup \big(\bigcup_{\{u,v\}\in \mathcal{E}}\,(\{(u,v)\}\times[0,W(u,v)])\big)/\sim$, where $\sim$ denotes the quotient defined by following identifications
\[
\begin{aligned}
    v & \sim ((v,u),0)  \quad\mbox{ for all } \{u,v\}\in \mathcal{E}\\
    ((v,u),t)& \sim ((u,v),W(\{u,v\})-t) \quad\mbox{ for all } \{u,v\}\in \mathcal{E}\mbox{ and all } t\in [0,W(\{u,v\})]
\end{aligned}
\]
and with metric $d_{X_{\mathcal{T}}}$ on $X_{\mathcal{T}}$ maps any pair of (equivalence classes) $((u_0,u_0),t)$ and $((v_0,v_1),s)$ in $X_{\mathcal{T}}$ to the non-negative number
\[
        \min_{i,j\in\{0,1\}}
        \,
        |t-i W(u_0,u_1)|
        +
        d_{\mathcal{T}}(u_i,v_j)
        +
        |s-jW(v_0,v_1)|
.
\]
\end{definition}

We call a metric space a \emph{simplicial tree} if it is essentially the same as a tree whose edges are finite closed real intervals, with the shortest path distance.
Simplicial trees are a special case of the following broader class of well-studied metric spaces, which we introduce to synchronize with the metric geometry literature, since it formulated many results using this broader class.  

\begin{definition}[$\mathbb{R}$-Tree]
A metric space $(X,d)$ is called an $\mathbb{R}$-tree if $X$ is connected
and for all $x,y,z,w\in X$,
$$(x,y)_w \ge \min \{ (x,z)_w, (z,y)_w \}$$
where $(x,y)_w$ denotes the Gromov product
$$(x,y)_w = \frac{1}{2}[d(x,w)+d(w,y)-d(x,y)].$$
\end{definition}

\begin{definition}[Valency]
The valency of the geometric realization of a metric space $X$ at a point $x$ is defined as the cardinality of the set of connected components of 
$X \backslash \{x\}$.
\end{definition}

\begin{proof}[{Proof of Lemma~\ref{lem:ConstructionEmbeddingAtInfinity}}]
\hfill\\
\textbf{Overview:}
\textit{The proof of this lemma can be broken down into $3$ steps.  First, we isometrically embed the tree into an $\mathbb{R}$-tree thus representing our discrete space as a more tractable connected (uniquely) geodesic metric space.  
This $\mathbb{R}$-tree is then isometrically embedded into a canonical $\mathbb{R}$-tree whose structure is regular and for which embeddings are exhibited more easily.  
Next, we ``asymptotically embed' this regular $\mathbb{R}$-tree into the boundary of the hyperbolic space, upon perturbing the embedding and adjusting the curvature of the hyperbolic space.  
We deduce the lemma upon composing all three embeddings.}
\hfill\\
\noindent\textit{Step 1 - Isometric Embedding of $(V,d_{\mathcal{T}})$ Into an $\mathbb{R}$-Tree}
\hfill\\

% \Anastasis{
If $V$ has only one point, then the result is trivial.  Therefore, we will always assume that $V$ has at least two points.  
\\
For each vertex $v\in V$, pick a different $w^v\in V$ such that $\{v,w^v\}\in \mathcal{E}$ and $W(v,w^v)\ge W(v,u)$ for all $\{u,v\}\in \mathcal{E}$; i.e.\ $w^v$ is adjacent to $v$ in the weighted graph $\mathcal{T}=(V,\mathcal{E},\mathcal{W})$.  
Consider the map $\varphi_1:V\rightarrow X_{\mathcal{T}}$ defined for any vertex $v\in V$ by
\[
        \varphi_1:
        v
    \mapsto 
        ((v,w^v),0)
.
\]
By definition $(X_{\mathcal{T}},d_{X_{\mathcal{T}}})$ is a simplicial tree and therefore by \citep[Corollary 3.1.13]{DasSimmonsUrbanski_2017_AMSBook_GeometryDynamicsGromovHyperbolicSpaces} it is an $\mathbb{R}$-tree.  In every $\mathbb{R}$-tree there is a unique shortest path (geodesic) connecting any pair of points.  This is because, by \citep[Observation 3.2.6]{DasSimmonsUrbanski_2017_AMSBook_GeometryDynamicsGromovHyperbolicSpaces}, all $\mathbb{R}$-trees satisfy the $\operatorname{CAT}(-1)$ condition, as defined in \citep[Definition II.1.1]{BrisonHaefliger_Book_1999__NPCMetricSpaces}, and in any metric spaces satisfying the $\operatorname{CAT}(-1)$ there is exactly one shortest path connecting every pair of points by \citep[Chapter II.1 - Proposition 1.4 (1)]{BrisonHaefliger_Book_1999__NPCMetricSpaces}.  Moreover, \citep[Chapter 3 - Lemma 1.4]{Chiswell_2001_Book__IntroductionToLambdaTrees} implies that if $x=x_0,\dots,x_N=y$ are (distinct) points in an $\mathbb{R}$-tree, for some $N\in \mathbb{N}$, such as $(X_{\mathcal{T}},d_{X_{\mathcal{T}}})$, for lying on \textit{the} geodesic (minimal length path) joining $x$ to $y$ then
\begin{equation}
\label{eq:setup_PathT_to_PathXT__setup}
        d_{X_{\mathcal{T}}}(x,y) 
    = 
        \sum_{i=0}^{N-1}\,
            d_{X_{\mathcal{T}}}(x_i,x_{i+1})
    .
\end{equation}
\\
Since $\mathcal{T}=(V,\mathcal{E},\mathcal{W})$ is a weighted tree, there is exactly one path joining any two nodes in a tree comprised of distinct points (a so-called reduced path in $\mathcal{T}$), independently of the weighting function $\mathcal{W}$, see e.g.~\citep[Chapter 2 - Lemma 1.4]{Chiswell_2001_Book__IntroductionToLambdaTrees}.  Therefore, for any $v,u\in V$ there exist one such unique finite sequence $u=u_1,\dots,u_N=v$ of distinct points (whenever $u\neq v$ with the case where $u=v$ being trivial).  By definition of $\varphi_1$ and the above remarks on $(X_{\mathcal{T}},d_{X_{\mathcal{T}}})$ being uniquely geodesic, we have that there exists exactly one  geodesic (minimal length curve) $\gamma:[0,1]\rightarrow X_{\mathcal{T}}$ satisfying
\[
    \gamma(t_i) = \varphi(u_i)
\]
for some distinct ``times'' $0=t_0<\dots<t_N=1$ with constant speed.  Therefore,~\eqref{eq:setup_PathT_to_PathXT__setup} implies that 
\begin{equation}
\label{eq:setup_PathT_to_PathXT__setup_B}
        d_{X_{\mathcal{T}}}(\varphi(u),\varphi(v)) 
    = 
        \sum_{i=0}^{N-1}\,
            d_{X_{\mathcal{T}}}(\varphi(u_i),\varphi(u_{i+1}))
    .
\end{equation}
Since, for $i=0,\dots,N-1$, $u_i$ and $u_{i+1}$ are adjacent in $\mathcal{T}$, meaning that $\{u_i,u_{i+1}\}\in\mathcal{E}$, then $d_{X_{\mathcal{T}}}(\varphi(u_i),\varphi(u_{i+1}))$ reduces to
\begin{align}
\label{eq:simple_form_distance_function__BEGIN}
    d_{X_{\mathcal{T}}}(\varphi(u_i),\varphi(u_{i+1}))
= &
    \underset{
        k=0,1,\,j=0,1
        }{\min}
    \,
        \big|
            0
            -
            k
            W(u_i,w^{u_i})
        \big|
\\
\notag
&   
    +
        d_{\mathcal{T}}(w_k,v_j)
\\
\notag
&   
    +
        \big|
            0
            -
            j
            W(u_{i+1},w^{u_{i+1}})
        \big|
\\
\notag
= &
    \underset{
        k=0,1,\,j=0,1
        }{\min}
    \,
        \big|
            0
            -
            0
            W(u_i,w^{u_i})
        \big|
\\
\label{eq:definition_shortestpath}
&   
    +
        W(u_i,u_{i+1})
\\
\notag
&   
    +
        \big|
            0
            -
            0
            W(u_{i+1},w^{u_{i+1}})
        \big|
\\
\label{eq:simple_form_distance_function__END}
= & W(u_i,u_{i+1})
,
\end{align}
where $w_0=u_i$, $w_1=w^{u_i}$, $v_0=u_{i+1}$, $v_1=w^{u_{i+1}}$ and~\eqref{eq:definition_shortestpath} holds by definition of $w^{u_i}$ and $w^{u_{i+1}}$ together with the fact that $\{u_i,u_{i+1}\}\in\mathcal{E}$ which implies that $\{u_i,u_{i+1}\}$ is a geodesic in $(V,d_{\mathcal{T}})$.  Combining the computation in~\eqref{eq:simple_form_distance_function__BEGIN}-\eqref{eq:simple_form_distance_function__END} with~\eqref{eq:setup_PathT_to_PathXT__setup_B} yields
\begin{equation}
\label{eq:isometry_1}
        d_{X_{\mathcal{T}}}(\varphi(u),\varphi(v)) 
    = 
        \sum_{i=0}^{N-1}\,
            W(u_i,u_{i+1})
    =
        d_{\mathcal{T}}(u,v)
,
\end{equation}
where the right-hand side of~\eqref{eq:isometry_1} holds since $(\{u_i,u_{i+1}\})_{i=0}^{N-1}$ was the unique path in $\mathcal{T}$ of distinct point from $u$ to $v$ and by definition of the shortest path distance in a graph.
Consequentially,~\eqref{eq:isometry_1} shows that $\varphi_1$ is an isometric embedding of $(V,d_{\mathcal{T}})$ into $(X_{\mathcal{T}},d_{X_{\mathcal{T}}})$.

\hfill\\
\noindent\textit{Step 2 - Embedding $(X_{\mathcal{T}},d_{X_{\mathcal{T}}})$ Into A Universal $\mathbb{R}$-Tree}
\hfill\\
Since $(X_{\mathcal{T}},d_{X_{\mathcal{T}}})$ has valency at-most $1\le \mu=\operatorname{deg}(\mathcal{T})<\#\mathbb{N}<2^{\aleph_0}$, then \citep[Theorem 1.2.3 (i)]{Dyubina_Polterovich_2001_BulLondMathSoc__ExplicitUniversalRTrees} implies that there exists\footnote{The metric space $(A_{\mu},d_{A_{\mu}})$ is constructed explicitly in \citep[Definition 1.1.1]{Dyubina_Polterovich_2001_BulLondMathSoc__ExplicitUniversalRTrees} but its existence dates back earlier to \cite{Mikel_1989_MemAmerMathSoc__UniversalRTrees,Mayer_Kikiel_Oversteegen_1992__TransAmerMathSoc__UniversalRTrees}.} an $\mathbb{R}$-tree $(A_{\mu},d_{A_{\mu}})$ of valency at-most $\mu$ and an isometric embedding $\varphi_2:(X_{\mathcal{T}},d_{X_{\mathcal{T}}})\rightarrow (A_{\mu},d_{A_{\mu}})$.

\hfill\\
\noindent\textit{Step 3 - The Universal $\mathbb{R}$-Tree At $\infty$ In the Hyperbolic Space $(\mathbb{H}_{-1}^d,d_{-1})$.}
\hfill\\
By \citep[Proposition 6.17]{BrisonHaefliger_Book_1999__NPCMetricSpaces} the hyperbolic spaces $(\mathbb{H}^d_{-1},d_{-1})$ have the structure of a simply connected and geodesically complete Riemannian manifold and by the computations on \citep[pages 276-277]{Jost_2017_Book__RiemannianGeometryGeometricAnalysis} it has constant negative sectional curvature equal to $-1$.  Now, since $\mu<2^{\aleph_0}$ then the just-discussed properties of $(\mathbb{H}^d_{-1},d_{-1})$ guarantee that \citep[Theorem 1.2.3 (i)]{Dyubina_Polterovich_2001_BulLondMathSoc__ExplicitUniversalRTrees}. 

Now \citep[Theorem 1.2.3 (i)]{Dyubina_Polterovich_2001_BulLondMathSoc__ExplicitUniversalRTrees}, together with \citep[Definition 1.2.1]{Dyubina_Polterovich_2001_BulLondMathSoc__ExplicitUniversalRTrees}, imply that the following holds: There is a diverging sequence $(\lambda_n)_{n=0}^{\infty}$ of positive real numbers such that for every $x\in A_{\mu}$ there is a sequence $(x^n)_{n=0}^{\infty}$ in $\mathbb{H}^d$ such that for every $\varepsilon>0$ there is an $n_{\varepsilon}\in \mathbb{N}_+$ such that for every integer $n\ge n_{\varepsilon}$ we have
\begin{equation}
\label{eq:embedding_at_infinity}
    \sup_{x,y\in A_{\mu}}
    \,
        \biggl|
                \frac{
                    d_{-1}(x^n,y^n)
                }{
                    \lambda_n
                }
            -
                d_{A_{\mu}}(x,y)
        \biggr|
    < 
        \varepsilon/2
.
\end{equation}
In particular,~\eqref{eq:embedding_at_infinity} holds for all $x,y\in \varphi_2\circ \varphi_1(V)\subseteq A_{\mu}$.  Since $V$ is finite, then so is $\varphi_2\circ \varphi_1(V)$ and since $\mathbb{H}^d$ is simply connected then for every $x\in \varphi_2\circ \varphi_1(V)$ there exists a point $\tilde{x}^{n_{\varepsilon}}$ for which $d_{-1}(x^{n_{\varepsilon}},\tilde{x}^{n_{\varepsilon}}) < \varepsilon/4$ and such that $\{\tilde{x}^{n_{\varepsilon}}\}_{x\in \varphi_2\circ\varphi_1(V)}$ and $\varphi_2\circ \varphi_1(V)$ have equal numbers of points.  Since $\varphi_2$ and $\varphi_1$ are isometric embeddings then they are injective; whence, $\{\tilde{x}^{n_{\varepsilon}}\}_{x\in \varphi_2\circ\varphi_1(V)}$ and $V$ have equal numbers of points.  Define $\varphi_3:\varphi_2\circ \varphi_1(V)\rightarrow \mathbb{H}^d$ by $x\mapsto \tilde{x}^{n_{\varepsilon}}$, for each $x\in \varphi_2\circ \varphi_1(V)$.

Therefore, the map $\varphi_3:\varphi_2\circ \varphi_1(V)\rightarrow \mathbb{H}^d$ is injective and, by~\eqref{eq:embedding_at_infinity}, it satisfies 
\[
    \max_{x,y\in \varphi_2\circ \varphi_1(V)}
    \,
        \biggl|
                \frac{
                    d_{-1}(\varphi_3(x),\varphi_3(y))
                }{
                    \lambda_n
                }
            -
                d_{A_{\mu}}(x,y)
        \biggr|
    < 
        \varepsilon
.
\]
Define the map $f^{\star}:\mathbb{R}^n\rightarrow \mathbb{H}^d$ as \textit{any} extension of the map $\varphi_3\circ \varphi_2\circ \varphi_1:V\rightarrow \mathbb{H}^d$.  Thus, $f^{\star}|_V=(\varphi_3\circ \varphi_2\circ \varphi_1)|_V$ and therefore $f^{\star}$ satisfies the following
\allowdisplaybreaks
\begin{align}
\label{eq:embedding_at_infinity__modified__BEING} 
            d_{\mathcal{T}}(u,v)
        -
            \varepsilon
    = &
            d_{A_{\mu}}(\varphi_2\circ \varphi_1(u),\varphi_2\circ \varphi_1(v))
        -
            \varepsilon
\\
\notag
    < & 
        \frac{
            d_{-1}(f^{\star}(u),f^{\star}(v))
        }{
            \lambda_n
        }
\\
\notag
    < & 
            d_{A_{\mu}}(\varphi_2\circ \varphi_1(u),\varphi_2\circ \varphi_1(v))
        +
            \varepsilon
\\
    = &
            d_{\mathcal{T}}(u,v)
        +
            \varepsilon
\label{eq:embedding_at_infinity__modified__END}
\end{align}
for each $u,v\in V$; where the equalities~\eqref{eq:embedding_at_infinity__modified__BEING} and~\eqref{eq:embedding_at_infinity__modified__END} held by virtues of $\varphi_2$ and $\varphi_1$ being isometries and since the compositions of isometries is itself an isometry.  

\hfill\\
\noindent\textit{Step 4 - Selecting The Correct Curvature on $\mathbb{H}^d_{\kappa}$ by Re-scaling The Metric $d_{-1}$.}
\hfill\\
Set $\kappa_{\varepsilon}\eqdef -\lambda_{n_{\varepsilon}}^2$.  By definition of $d_{\kappa_{\varepsilon}}$, see \citep[Definition 2.10]{BrisonHaefliger_Book_1999__NPCMetricSpaces}, the chain of inequalities in~\eqref{eq:embedding_at_infinity__modified__BEING}-\eqref{eq:embedding_at_infinity__modified__END} imply that
\begin{equation}
\label{eq:embedding_at_infinity__RoughIsometryFormOverFiniteSet}
        d_{\mathcal{T}}(u,v)
    -
        \varepsilon
    <
        d_{\kappa_{\varepsilon}}(f^{\star}(u),f^{\star}(v))
    <
        d_{\mathcal{T}}(u,v)
    +
        \varepsilon
.
\end{equation}
Set $\delta \eqdef \min_{x,\Tilde{x}\in V; x\ne\Tilde{x}} d_{\mathcal{T}}(x,\Tilde{x})>0$ since $V$ is finite.  Note that for any $0<\varepsilon<\delta$, the distortion of $f^{\star}$ is at most
\begin{equation}
\label{eq:embedding_at_infinity__RelationOfRoughIsometryToDistortion}
        \max_{\substack{x,\Tilde{x}\in V \\ x\ne\Tilde{x}}} 
        \,
            \frac{d_{\mathcal{T}}(x,\Tilde{x})+\varepsilon}{d_{\mathcal{T}}(x,\Tilde{x})-\varepsilon}
    =
        \frac{\delta+\varepsilon}{\delta-\varepsilon}
\end{equation}
and that the right-hand side of~\eqref{eq:embedding_at_infinity__RelationOfRoughIsometryToDistortion} tends to $1$ as $\varepsilon\rightarrow 0$.  Thus, we may choose $\varepsilon>0$ small enough to ensure that~\eqref{eq:embedding_at_infinity__RoughIsometryFormOverFiniteSet} holds; relabelling $\kappa\eqdef \kappa_{\varepsilon}$ accordingly.
\end{proof}

\section{Conclusion}
\label{s:Conclusion}
We have established lower bounds on the smallest achievable distortion by any Multi-Layer Perceptron (MLP) embedding a large latent metric tree into a Euclidean space, as proven in Theorem~\ref{thm:LowerBound}.  Our lower bound holds true independently of the depth, width, number of trainable parameters, and even the (possibly discontinuous) activation function used to define the MLP.

In contrast to this lower bound, we have demonstrated that Hyperbolic Neural Networks (HNNs) can effectively represent any latent tree in a $2$-dimensional hyperbolic space, with a trainable constant curvature parameter. Furthermore, we have derived upper bounds on the capacity of the HNNs implementing such an embedding and have shown that it depends at worst polynomially on the number of nodes in the graph.

To the best of the authors' knowledge, this constitutes the initial proof that HNNs are well-suited for representing graph structures, while also being the first evidence that MLPs are not. Thus, our results provide mathematical support for the notion that HNNs possess a superior inductive bias for representation learning in data with latent hierarchies, thereby reinforcing a widespread belief in the field of geometric deep learning.

\section{Acknowledgment and Funding}
\label{s:AcknowledgmentandFunding}
AK acknowledges financial support from the NSERC Discovery Grant No.\ RGPIN-2023-04482 and their McMaster Startup Funds.  RH was funded by the James Steward Research Award and by AK's McMaster Startup Funds.  HSOB acknowledges financial support from the Oxford-Man Institute of Quantitative Finance for computing support
The authors also would like to thank Paul McNicholas and A.\ Martina Neuman their for their helpful discussions.

\bibliography{Bookeaping/reference}
\end{document}